\documentclass[11pt, a4paper,leqno]{amsart}
\usepackage{amsmath,amsthm,amscd,amssymb,amsfonts, amsbsy}
\usepackage{latexsym}
\usepackage{txfonts}
\usepackage{exscale}

\usepackage[colorlinks,citecolor=red,pagebackref,hypertexnames=false]{hyperref}

\usepackage[colorinlistoftodos,textwidth=20mm]{todonotes}
\usepackage{dsfont}

\usepackage{pgf}

\usepackage[font=scriptsize,labelfont=bf]{caption}
\usepackage[font=scriptsize,labelfont=bf]{subcaption}
\usepackage{color,soul}

\usepackage{placeins}

\usepackage[export]{adjustbox}

\calclayout
\allowdisplaybreaks

\theoremstyle{definition}

\numberwithin{equation}{section}
\def\barint{\,\Xint -} 
\def\bariint{\barint_{} \kern-.4em \barint}
\def\bariiint{\bariint_{} \kern-.4em \barint}

\newcommand{\beq}{\begin{equation}}
\newcommand{\bea}[1]{\begin{array}{#1} }
\newcommand{\eeq}{ \end{equation}}
\newcommand{\ea}{ \end{array}}

\def \y {y}

\def\mean#1{\mathchoice%
          {\mathop{\kern 0.2em\vrule width 0.6em height 0.69678ex depth -0.58065ex
                  \kern -0.8em \intop}\nolimits_{\kern -0.4em#1}}%
          {\mathop{\kern 0.1em\vrule width 0.5em height 0.69678ex depth -0.60387ex
                  \kern -0.6em \intop}\nolimits_{#1}}%
          {\mathop{\kern 0.1em\vrule width 0.5em height 0.69678ex
              depth -0.60387ex
                  \kern -0.6em \intop}\nolimits_{#1}}%
          {\mathop{\kern 0.1em\vrule width 0.5em height 0.69678ex depth -0.60387ex
                  \kern -0.6em \intop}\nolimits_{#1}}}

\def\vintslides_#1{\mathchoice%
          {\mathop{\kern 0.1em\vrule width 0.5em height 0.697ex depth -0.581ex
                  \kern -0.6em \intop}\nolimits_{\kern -0.4em#1}}%
          {\mathop{\kern 0.1em\vrule width 0.3em height 0.697ex depth -0.604ex
                  \kern -0.4em \intop}\nolimits_{#1}}%
          {\mathop{\kern 0.1em\vrule width 0.3em height 0.697ex depth -0.604ex
                  \kern -0.4em \intop}\nolimits_{#1}}%
          {\mathop{\kern 0.1em\vrule width 0.3em height 0.697ex depth -0.604ex
                  \kern -0.4em \intop}\nolimits_{#1}}}

\newcommand{\aveint}[2]{\mathchoice%
          {\mathop{\kern 0.2em\vrule width 0.6em height 0.69678ex depth -0.58065ex
                  \kern -0.8em \intop}\nolimits_{\kern -0.45em#1}^{#2}}%
          {\mathop{\kern 0.1em\vrule width 0.5em height 0.69678ex depth -0.60387ex
                  \kern -0.6em \intop}\nolimits_{#1}^{#2}}%
          {\mathop{\kern 0.1em\vrule width 0.5em height 0.69678ex depth -0.60387ex
                  \kern -0.6em \intop}\nolimits_{#1}^{#2}}%
          {\mathop{\kern 0.1em\vrule width 0.5em height 0.69678ex depth -0.60387ex
                  \kern -0.6em \intop}\nolimits_{#1}^{#2}}}

\def\eqn#1$$#2$${\begin{equation}\label#1#2\end{equation}}
\def\charfn_#1{{\raise1.2pt\hbox{$\chi
_{\kern-1pt\lower3pt\hbox{{$\scriptstyle#1$}}}$}}}

\def\qq1{q_*}
\def\q2{q_{**}}

\newcommand{\ve}{\varepsilon}

\newdimen\vintbar
\def\vint{-\kern-\vintbar\int}

\newcommand{\rset}{\mathbb{R}}

\def\0{\boldsymbol 0}

\DeclareMathOperator*{\argmin}{arg\min}

\newcommand{\N}{\mathbb N}

\newtoks\by
\newtoks\paper
\newtoks\book
\newtoks\jour
\newtoks\yr
\newtoks\pages
\newtoks\vol
\newtoks\publ

\def\name[#1, #2]{#1 #2}
\def\ota{{\hbox{\bf ???}}}
\def\cLear{\by=\ota\paper=\ota\book=\ota\jour=\ota\yr=\ota
\pages=\ota\vol=\ota\publ=\ota}
\def\endpaper{\the\by, \textit{\the\paper},
{\the\jour} \textbf{\the\vol} (\the\yr), \the\pages.\cLear}
\def\endbook{\the\by, \textit{\the\book},
\the\publ, \the\yr.\cLear}
\def\endpap{\the\by, \textit{\the\paper}, \the\jour.\cLear}
\def\endproc{\the\by, \textit{\the\paper}, \the\book, \the\publ,
\the\yr, \the\pages.\cLear}

\begin{document}
\title[Deep learning, stochastic gradient descent and diffusion maps]{Deep learning, stochastic gradient\\ descent and diffusion maps}

\address{Carmina Fjellstr{\"o}m\\Department of Mathematics, Uppsala University\\
S-751 06 Uppsala, Sweden}
\email{carmina.fjellstrom@math.uu.se}

\address{Kaj Nystr\"{o}m\\Department of Mathematics, Uppsala University\\
S-751 06 Uppsala, Sweden}
\email{kaj.nystrom@math.uu.se}

\author{Carmina Fjellstr{\"o}m and  Kaj Nystr{\"o}m}
\maketitle
\begin{abstract}
\noindent
Stochastic gradient descent (SGD)  is widely used in deep learning due to its computational efficiency, but a complete understanding of why SGD performs so well remains a major challenge. It has been observed empirically that most eigenvalues of the Hessian of the loss functions on the loss landscape of over-parametrized deep neural networks are close to zero, while only a small number of eigenvalues are large. Zero eigenvalues indicate zero diffusion along the corresponding directions. This indicates that the process of minima selection mainly happens in the relatively low-dimensional subspace corresponding to the top eigenvalues of the Hessian. Although the parameter space is very high-dimensional, these findings seems to indicate that the SGD dynamics may mainly live on a low-dimensional manifold.  In this paper, we pursue a truly data driven approach to the problem of getting a potentially deeper understanding of the high-dimensional parameter surface, and in particular, of the landscape traced out by SGD by analyzing the data generated through SGD, or any other optimizer for that matter, in order to possibly discover (local) low-dimensional representations of the optimization landscape.  As our vehicle for the exploration, we use diffusion maps introduced by R. Coifman and coauthors.\\

\noindent
2000  {\em Mathematics Subject Classification.}
\noindent

\medskip

\end{abstract}



    \setcounter{equation}{0} \setcounter{theorem}{0}
    \section{Introduction and motivation}

    The calibration of deep neural networks results in the optimization problem
\begin{align}\label{optprob}
\mathbf{x}^\star = \argmin_{\mathbf{x} \in \rset^m} \Bigl\{ f(\mathbf{x}) := \frac1{N} \sum\nolimits_{i=1}^N f_{i}(\mathbf{x}) \Bigr\},
\end{align}
where $\mathbf{x}\in\rset^m$ denotes the weights of the neural network and $f:\rset^m\to\rset$ is the loss function, which typically is non-convex as a function of  $\mathbf{x}$. $f_{i}$, for $i \in \{1, \dots, N\}$,  denotes the contribution to the loss function from data point $i$ and  $N$ denotes the total number of data points.

A natural approach to the optimization problem in \eqref{optprob} is to use gradient descent (GD).  However, when $N$ is large, it may be computationally prohibitive to compute the full gradient of the objective function $f$ and so stochastic gradient descent (SGD) provides an alternative. SGD is based on a (noisy) gradient evaluated from a single data point or a minibatch of data points, resulting in the iterative updates
\begin{align}
 \mathbf{x}(t_{j+1})=\mathbf{x}_{j+1}= \mathbf{x}_{j} - \eta \nabla \widetilde{f}^{(j)}(\mathbf{x}_j)=\mathbf{x}(t_{j}) - \eta \nabla \widetilde{f}^{(j)}(\mathbf{x}(t_j)),\, t_{j+1}=t_j+\eta, \label{eqn:sgd_main}
\end{align}
where $j \in \{0, \dots, M\}$ denotes the iteration number, and $\nabla \widetilde{f}^{(j)}$ denotes the stochastic gradient at iteration $j$ defined as
\begin{align}
\label{eqn:stoch_grad}
\nabla \widetilde{f}^{(j)} (\mathbf{x}):= \frac1{n_j} \sum\nolimits_{i \in \Omega_j}  \nabla f_{i}(\mathbf{x}).
\end{align}
Here, $\Omega_j \subset \{1,\dots,N\}$ is a random subset that is drawn with or without replacement at iteration $j$, and $n_j$ denotes the number of elements in $\Omega_j$. When no confusion arises, we simply write $\Omega$ and $n$. The $\eta>0$ in \eqref{eqn:sgd_main}, which  can either be constant or varying with the iteration, is known as the learning rate.

Given the use of SGD, a set or sequence of points $X:=\{\mathbf x_j\}_{j=1}^M=\{\mathbf{x}(t_j)\}_{j=1}^M$ is generated, either from one sequence of runs of SGD or merged from several different runs. In particular, the set $X$ contains the information in the paths of the SGD  in the  high-dimensional space of parameters $\mathbb R^m$. In general, it is difficult to picture the geometry of the loss surface $$\Sigma:=\{(\mathbf{x},f(\mathbf{x})):\ \mathbf{x}\in\mathbb R^m\}$$ and insightful descriptions of this loss landscape as well as the geometry traced out by the paths of the SGD is still lacking due to the fact that while $f$ may be smooth, it is a non-linear, non-convex function in $\mathbb R^m$ with $m$ truly large. Heuristically, one way to think of the loss surface $\Sigma$, though the picture seems to be even more complex in reality, is as a landscape with peaks and valleys separated by ridges. Therefore, any optimizer including SGD could potentially get trapped in a basin and valley enclosing a local minima, finding it difficult to move from its initialized value over the ridges in the direction of the global minima.

The loss landscape or loss surface $\Sigma$ has received a lot of attention in the literature.  To mention a few relevant papers, \cite{ choromanska2015loss,dauphin2014identifying} conjectured that local minima of multi-layer neural networks have similar loss function values, and proved the result in idealized settings. For linear networks, it is known  \cite{kawaguchi2016deep} that all local minima are also globally optimal. Several theoretical works have explored whether a neural network has spurious valleys (non-global minima that are surrounded by other points with higher loss). \cite{freeman2016topology} showed that for a two-layer network, if it is sufficiently over-parametrized, then all the local minimizers are (approximately) connected. However, in order to guarantee a small loss along the path, they need the number of neurons to be exponential in the number of input dimensions. \cite{venturi2018spurious} proved that if the number of neurons is larger than either the number of training samples or the intrinsic dimension, then the neural network cannot have spurious valleys.  \cite{liang2018understanding} proved similar results for the binary classification setting. We also refer to \cite{freeman2016topology, garipov2018loss, liang2018understanding, nguyen2019connected, nguyen2018loss, venturi2018spurious} for insightful discussions concerning the geometry of the loss landscape.

SGD is widely used in deep learning due to its computational efficiency, but understanding how SGD performs better than its full batch counterpart in terms of test accuracy remains a major challenge. While  SGD seems to find zero loss solutions on the loss landscape $\Sigma$, at least in certain regimes, it appears that the algorithm finds solutions with different properties depending on how it is tuned, and a satisfactory theory explaining the success of SGD is in several ways  still lacking. Empirically, it has been observed that SGD can usually find flat minima among a large number of sharp minima and local minima \cite{hochreiter1995simplifying,hochreiter1997flat}. Other papers indicate that learning flat minima is closely related to the problem of generalization \cite{dinh2017sharp, dziugaite2017computing, hardt2016train, kleinberg2018alternative,   hoffer2017train,  neyshabur2017exploring, wu2017towards, zhang2017understanding}. Several papers are also devoted to flatness itself,  measuring flatness \cite{hochreiter1997flat, sagun2017empirical, yao2018hessian}, rescaling flatness \cite{tsuzuku2019normalized, xie2020stable}, and finding flatter minima \cite{chaudhari2017entropy, NIPS2019_8524, hoffer2017train,  xie2020artificial}. Furthermore, it has been observed that most eigenvalues of the Hessian at the loss landscape of over-parameterized deep neural networks are close to zero, and in particular, only a small number of eigenvalues are large \cite{li2018visualizing, sagun2017empirical}. Zero eigenvalues indicate zero diffusion along the corresponding directions and, theoretically, one may be inclined to ignore these zero-eigenvalue directions.  A small number of large eigenvalues means that the process of minima selection mainly happens in the relatively low-dimensional subspace corresponding to the top eigenvalues of the Hessian \cite{gur2018gradient}. In particular, although the parameter space is high-dimensional, SGD dynamics depends only modestly on the dimensions corresponding to small second-order directional derivatives, and SGD can heuristically be pictured as exploring the parameter space around a minimum in a much lower dimensional space. Still, a quantitative theory explaining  these phenomena is lacking.

In this paper, we pursue a truly data driven approach with the ambition to contribute to the  understanding of the loss landscape $\Sigma$, and in particular to the  understanding of the landscape traced out by SGD, by analyzing the data generated through SGD, or any other optimizer for that matter, in order to possibly discover (local) low-dimensional representations of $\Sigma$ and the optimization landscape. Note that this discovery of low-dimensional representations of high-dimensional data, characterization of the underlying geometry, and description of the density are some of the fundamental problems in data science. In general, to achieve this, statistical tools are used on SGD paths to detect the slow variables, meta-stable states, as well as connections and transition times between these states. This is the focus of this paper as we explore a low-dimensional representation of the high-dimensional data $X$ generated by SGD. As the vehicle for our exploration, we use the insightful work of R. Coifman and collaborators on diffusion maps and geometry and the relation to Langevin dynamics and Fokker-Planck equations.

Diffusion maps and geometry  \cite{Coifman2006, Lafon2006} are tools for the analysis of large datasets. A family of random walk processes on the large data set is constructed using isotropic and anisotropic diffusion kernels. Afterwhich, the eigenvalues and eigenvectors are analyzed, where the most dominant ones are known to be the principal components. These principal components contain key information regarding the geometry and statistics of the underlying space. Today, diffusion maps, based on the construction of the graph Laplacian of the data set \cite{Coifman2006a}, is an established manifold learning technique that has found application in many areas including  signal processing, image processing and machine learning  \cite{Coifman:2014, David2012, Farbman:2010, Gepshtein:2013, Haddad2014, Lafon2006, Mishne2013, Singer:2009, Talmon2012}.

The theme in our paper and in the works of R. Coifman and collaborators, see \cite{Coifman2006, Lafon2006, Talmon2012}, for example, is that while many dynamical systems initially may seem to require  high-dimensional spaces, coarser length and time scales normally reveal an intrinsic low dimensionality. Often, this low dimensionality can be captured by only a few variables known as the reaction coordinates. Dimension reduction as well as the derivation of complex operators based on which such systems under coarser scales evolve are, therefore, central undertakings.

\subsection{Organization of the paper} The paper is organized as follows. In Section \ref{sec:diff_maps}, we introduce the necessary background concerning diffusion maps and geometry, kernels, the Mahalanobis distance and SGD. In Section \ref{emp}, the most extensive  part of the paper, we analyze the high-dimensional parameter surface in the context of two different neural network architectures and two different data sets. The two data sets used are the iris flower for a classification problem and the auto miles per gallon (MPG) for a regression problem. In Section \ref{sumu}, we summarize our results, state conclusions and discuss directions for future research.

\section{Diffusion maps and geometry}\label{sec:diff_maps}
In the following, we let $X:=\{\mathbf x_i\}_{i=1}^M=\{\mathbf{x}(t_i)\}_{i=1}^M$, where $\mathbf x_i=\mathbf{x}(t_i)\in\mathbb R^m$. We stress that the integers $(N,n,M,m)$ refer to the number of samples $(N)$ used in the definition of the loss function $f$, the number of samples $(n)$ used in the calculation of the gradient in SGD, the number of samples $(M)$ in the path(s) generated by SGD, and the dimension $(m)$ of the parameter space. These as well as other notations used throughout the paper are summarized in Table \ref{table:notation}.
{\renewcommand{\arraystretch}{1.3}
\begin{table}[hbt!]
\centering
\begin{tabular}{l|p{0.55\linewidth}}
\hline
Notation        & Description       \\\hline
$N$	&	number of samples			\\ 
$n$	&	batch size			\\ 
$M$	&	number SGD steps			\\ 
$m$	&	dimension of parameter space			\\ 
$\mathbf{x}\in\mathbb R^m$	&	model parameters, i.e., weights of the neural network\\	
$f(\mathbf{x}):\rset^m\to\rset$	&	loss function			\\
$\nabla f(\mathbf{x})$	&	full gradient			\\
$\nabla \widetilde{f} (\mathbf{x})$	& stochastic gradient				\\
$C(\mathbf{x})$ 	& covariance at point $\mathbf{x}$			\\
$\varepsilon$	& diffusion map parameter			\\
$d$	& dimension of lower dimensional subspace			\\
$\lambda_j, j=0, \ldots, M-1$	& eigenvalues obtained through diffusion mapping			\\
$\Lambda_i:=\frac{\sqrt{\lambda_1^2+\lambda_2^2+\dots+\lambda_i^2}}{\sqrt{\lambda_1^2+\lambda_2^2+\dots+\lambda_N^2}}$	& energy ratio of the eigenvalues		\\

\hline
\end{tabular}
\caption{Notation table.}
\label{table:notation}
\end{table}}

\subsection{The Mahalanobis distance and SGD}  Recall that if two points $\mathbf z(t_1)$ and $\mathbf z(t_2)$ are drawn from an $m$-dimensional
Gaussian distribution with covariance ${C}_z$, then  the Mahalanobis distance between the points, see \cite{mahalanobis1936generalized}, is defined as
\begin{equation}
	\| \mathbf z(t_1) - \mathbf z(t_2) \| _{MD} = \sqrt{ (\mathbf z(t_1) - \mathbf z(t_2))^\ast {C}_z^{-1} (\mathbf z(t_1) - \mathbf z(t_2) )  },
\end{equation}
where $^\ast$ denotes the transpose. In particular, if ${C}_z^{-1} = \mathrm{diag}(\sigma_1^{-1}, \ldots,  \sigma_m^{-1})$ is a constant matrix, then
\begin{equation} \label{eq:rescale_x_dist}
\| \mathbf z(t_1) - \mathbf z(t_2) \|^2_{MD} = \sum_{i=1}^m \sigma_i^{-1} \left( z_i(t_2) - z_i(t_1) \right)^2,
\end{equation}
where $z_i(\cdot)$ denotes the $i$-th coordinate of the vector $\mathbf z(\cdot)$.  Note that in \eqref{eq:rescale_x_dist}, the coordinates with large volatilities or standard deviations, determined by $\sigma_i$, make negligible contributions to the Mahalanobis distance, and these coordinates or variables may be referred to as the \textit{fast variables}. In particular, the  metric can be seen as  implicitly insensitive, or only modestly sensitive,  to changes in the fast variables. Introducing
\begin{equation} \label{eq:general_rescale}
y_i(t):= \frac 1{\sqrt{\sigma_i}} z_i(t),
\end{equation}
the metric \eqref{eq:rescale_x_dist} can be rewritten as
\begin{equation} \label{eq:norm_z}
\| {\mathbf z}(t_2) - \mathbf z(t_1) \|^2_{MD} = \|\mathbf y(t_2) - \mathbf y(t_1) \|^2_2.
\end{equation}
Using this notation, $\mathbf y(t)$ is a  stochastic process, rescaled so that each variable has unit diffusivity, with the same dimensionality as $\mathbf z(t)$. By performing this  rescaling, the problem is transformed from a problem of detecting the \textit{slow variables} within dynamic data, to a problem of more traditional data mining. In particular, by construction, the Mahalanobis distance takes into account information about the dynamics and relevant time scales, enabling the use of traditional data mining techniques when used  with this metric to detect the slow variables in the data \cite{singer2009detecting}.

Note that the  traditional Mahalanobis distance is defined for a fixed distribution,
whereas we are dealing with a distribution that possibly changes as a function of position due to
nonlinearities in the drift term of the SGD. To account for this, $\Vert\cdot\Vert_{MD}$ for us will denote a Mahalanobis distance calculated on vectors in $\mathbb R^m$ and engineered based on the (implicit) covariance structure of SGD. Indeed, given $\mathbf{x}(t_1), \mathbf{x}(t_2)\in \mathbb R^m$,  we use, see \cite{Coifman2016}, the modified Mahalanobis distance
\begin{equation}
\label{eq:mahalanobis_modified}
\Vert \mathbf{x}(t_1)-\mathbf{x}(t_2)\Vert_{MD}^2 = {\frac{1}{2}}{(\mathbf{x}(t_1)-\mathbf{x}(t_2))^\ast}{\big(C^\dagger(\mathbf{x}(t_1))+C^\dagger(\mathbf{x}(t_2))\big)}{(\mathbf{x}(t_1)-\mathbf{x}(t_2))},
\end{equation}
where $C(\mathbf x(t_j))$ is the covariance at the position/point $\mathbf x(t_j)$ and $\dagger$ denotes the Moore-Penrose pseudoinverse.

The SGD covariance at $\mathbf{x}$, where $\mathbf{x}$ is the model parameters, can be expressed as, see, for example, \cite{hoffer2017train, hu2019diffusion, smith2017bayesian,  wu2020noisy, xie2021covariance},
\begin{equation}
\label{eq:sgd_cov_exact}
C(\mathbf{x}) = \frac{N-n}{n(N-1)}{\Bigg[{\frac{1}{N}}{\sum_{i=1}^N \nabla_x f_i(\mathbf{x}) \nabla_x f_i(\mathbf{x})^\ast} -  \nabla_x f(\mathbf{x}) \nabla_x f(\mathbf{x})^\ast \Bigg]},
\end{equation}
the proof of which is detailed in Appendix \ref{appendix:proof_cov}. When $N$ is large, it may not be feasible to compute the full gradient for every iteration. Moreover, it has been observed, see, for example, \cite{wu2020noisy, xie2021covariance}, that near critical points, the second term in \eqref{eq:sgd_cov_exact} is dominated by the first. Hence, for final stages of optimization, the covariances can be approximated by 
\begin{equation}
\label{eq:sgd_cov_approx}
C(\mathbf{x}) \approx \frac{N-n}{n(N-1)}\frac{1}{N}{\sum_{i=1}^N \nabla_x f_i(\mathbf{x}) \nabla_x f_i(\mathbf{x})^\ast}.
\end{equation}

Let \begin{equation*}
\label{eq:sgd_cov_fisher}
F(\mathbf{x}):=\frac{1}{N}{\sum_{i=1}^N \nabla_x f_i(\mathbf{x}) \nabla_x f_i(\mathbf{x})^\ast}
\end{equation*}
be the Fisher information matrix. For $N \gg n$, $\frac{N-n}{N-1} \approx 1$, the approximation in \eqref{eq:sgd_cov_approx} simplifies even further and one can obtain, see \cite{zhu2018anisotropic}, that the SGD covariance is approximately proportional to the Hessian $H(\mathbf{x})$ of the loss function
\begin{equation}
\label{eq:sgd_cov_hessian}
C(\mathbf{x}) \approx \frac{1}{n}F(\mathbf{x}) \approx \frac{1}{n}{H(\mathbf{x})}.
\end{equation}

In practice, the covariance matrix can also be estimated from a short trajectory of samples in time around the sample $\mathbf{x}(t)$ by
\begin{equation}
	\widehat{{C}}(\mathbf{x}(t)) = \sum \limits _{\tau = t-L}^{t+L} (\mathbf{x}(\tau) - {\mu}(t))(\mathbf{x}(\tau) - \mu(t))^\ast,
	\label{eq:cov}
\end{equation}
where $\mu(t)$ is the empirical mean of the short trajectory of samples, and $2L$ is the length of the trajectory.

\subsection{Diffusion kernels and maps}\label{sec:diffusion_kernelmap} The starting point for the construction of the diffusion maps is a symmetric and non-negative kernel $k=k(\cdot,\cdot):\mathbb R^m\times\mathbb R^m\to \mathbb R$.  While many kernels satisfy this property, we will, in this paper, mainly use the kernel
\beq
\label{k_epsilon}
k(\mathbf{x}_i,\mathbf{x}_j):=k_\varepsilon(\mathbf{x}_i,\mathbf{x}_j):=\exp\left(-\Vert \mathbf{x}_i-\mathbf{x}_j\Vert^2/\varepsilon\right).
\eeq
Here, $\varepsilon>0$ is a global scale parameter, a degree of freedom, and $\Vert\cdot\Vert$ could, in principle, be any relevant distance function. For us, $\Vert\cdot\Vert$ will denote the Mahalanobis distance $\Vert\cdot\Vert_{MD}$ introduced in \eqref{eq:mahalanobis_modified}, calculated on vectors in $\mathbb R^m$  and engineered based on the (implicit) covariance structure of SGD.

 Given the  set $X:=\{\mathbf{x}_i\}_{i=1}^M=\{\mathbf{x}(t_i)\}_{i=1}^M$ of data points, we construct a weighted graph  with the data points as nodes.  Given the edge connecting two nodes $\mathbf{x}_i, \mathbf{x}_j \in X$, we let the weight of the edge be equal to $k(\mathbf{x}_i,\mathbf{x}_j)=k_\ve(\mathbf{x}_i,\mathbf{x}_j)$. In this context, $k(\mathbf{x}_i,\mathbf{x}_j)$ should be seen as a measure of similarity between the data points $\mathbf{x}_i, \mathbf{x}_j \in X$. Based on $X$ and $k$, we introduce a $M\times M$ dimensional matrix $K$ with entries ${K}[i,j] = {K}_\varepsilon[i,j] :=k(\mathbf{x}_i,\mathbf{x}_j)$.  In practice, ${K}$ can often  be computed using only the nearest neighbors of every point. In this case, ${K}[i,j]$ is defined to be zero for every $\mathbf{x}_j$ which is not among the nearest neighbors of $\mathbf{x}_i$. Naturally, a notion of nearest neighbors then has to be defined.

 To construct an approximation of the Laplace-Beltrami operator on the data set,  we first use a normalization of the data set; this is a natural preprocessing step and is necessary to ensure that the embeddings to be constructed do not rely on the distribution of the points~\cite{Coifman2006,Lafon2006}. Let $D$ be a $M\times M$ dimensional diagonal matrix with ${D}[i,i]:=\sum_{\mathbf{x}_j\in X}k(\mathbf{x}_i,\mathbf{x}_j)$. We then introduce a normalized  matrix ${\widetilde{K}}$ with entries ${\widetilde{K}}[i,j]$,
\begin{equation*}
\label{eq:ln_norm}
{\widetilde{K}} = {D}^{-1/2}{K}{D}^{-1/2}.
\end{equation*}
 Based on ${\widetilde{K}}$  we also introduce
\begin{equation}
\label{eq:random_walk}
{P} := {\widetilde{D}}^{-1}{\widetilde{K}}, \;\;\; {\widetilde{D}}[i,i]:=\sum_{j}{\widetilde{K}}[i,j].
\end{equation}
The row-stochastic matrix ${P}$ satisfies ${P}[i,j]\geq0$ and $\sum_{j}{P}[i,j]=1$ and, therefore, can be viewed as the transition matrix of a Markov chain on the data set $X$. ${P}$  has a sequence of biorthogonal left and right eigenvectors, $\phi_\ell$ and $\psi_\ell$, respectively, and a sequence of positive eigenvalues $\{\lambda_j\}_{j=0}^{M-1}$ satisfying $1 = |\lambda_0|\geq|\lambda_1|\geq ...$. Using this notation and introducing
\begin{equation}
\label{eq:eigen_decompose}
p_\tau(\mathbf{x}_i,\mathbf{x}_j):=\sum_{\ell\geq 0} \lambda^\tau_\ell\psi_\ell(\mathbf{x}_i)\phi_\ell(\mathbf{x}_j),\ \tau\geq 0,
\end{equation}
we can interpret $p_\tau(\mathbf{x}_i,\mathbf{x}_j)$ as the probability that the Markov chain, starting at $\mathbf{x}_i$ at $\tau=0$, is at $\mathbf{x}_j$ after $\tau$ steps.

We introduce  a distance $d(\mathbf{x}_i,\mathbf{x}_j,\tau)$ between two points $\mathbf{x}_i,\mathbf{x}_j \in X$,
\begin{equation}
\label{eq:diffusion_distance1}
d(\mathbf{x}_i,\mathbf{x}_j,\tau ) = \sum_{\mathbf{x}_k\in X}\frac{\big(p_\tau(\mathbf{x}_i,\mathbf{x}_k)-p_\tau(\mathbf{x}_j,\mathbf{x}_k)\big)^2}{\phi_0(\mathbf{x}_k)} = \sum_{\ell\geq1} \lambda^{2\tau}_\ell(\psi_\ell(\mathbf{x}_i)-\psi_\ell(\mathbf{x}_j))^2.
\end{equation}
Here $\phi_0$ denotes the stationary probability distribution on the graph. $d(\mathbf{x}_i,\mathbf{x}_j,\tau )$ is referred to as the diffusion distance between $\mathbf{x}_i$ and $\mathbf{x}_j$ at step/time $\tau$. The distance function/metric constructed is robust to noise, as the distance between any two points is a function of all possible paths of length $\tau$ between the points. The diffusion distance can, as a consequence of the decay of the spectrum of ${P}$, be approximated using only the first, say $d$, eigenvectors. Furthermore, as a consequence of \eqref{eq:diffusion_distance1}, a mapping  between the original space and the eigenvectors $\psi_\ell$ can be defined. Indeed, if one only keeps the first $d$ eigenvectors,  then the data set $X$ gets embedded into the Euclidean space $\mathbb{R}^{d}$ through the map $\Psi_\tau$. In this embedding, the diffusion distance is equal to the Euclidean distance:
\begin{equation}
\label{eq:diffusion_map}
\Psi_\tau:\mathbf{x}_i\rightarrow \big( \lambda_1^\tau\psi_1(\mathbf{x}_i), \lambda_2^\tau\psi_2(\mathbf{x}_i),..., \lambda_{d}^\tau\psi_{d}(\mathbf{x}_i)\big)^\ast.
\end{equation}
As $\psi_0$ is a constant vector, $\psi_0$ is not used in \eqref{eq:diffusion_map}.


\section{Empirical Investigation}\label{emp}

\begin{figure}[!htbp]
\centering
\includegraphics[width=0.6\textwidth]{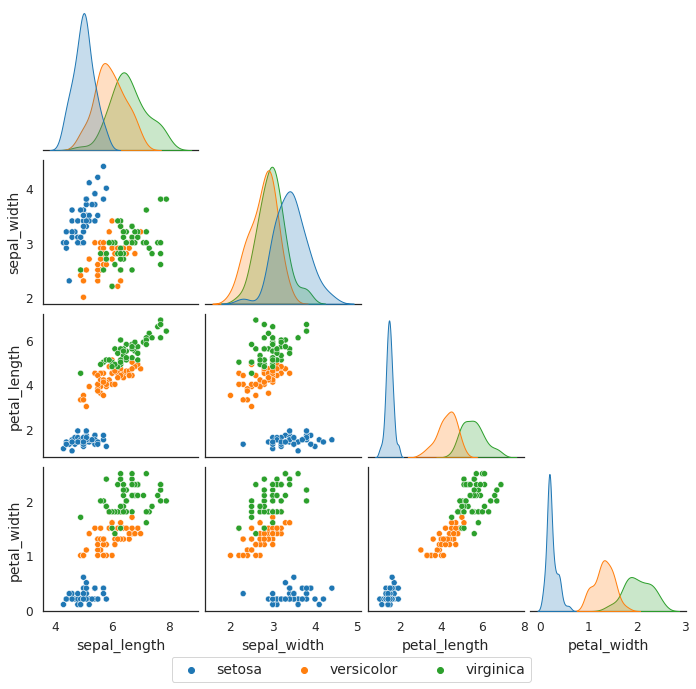}
\caption{Iris data set. Four features: petal length, petal width, sepal length, and sepal width. Three classes: setosa, versicolor, and virginica. Off-diagonal graphs are scatterplots of all samples based on pairwise combinations of the features shown in the x- and y-axes. Diagonal graphs are density estimates of the three classes for the particular feature shown in the x-axis.}
\label{fig:iris}
\end{figure}

To analyze the high-dimensional parameter surface in practice using the diffusion maps discussed in Section \ref{sec:diff_maps}, and based on the distance function in \eqref{eq:mahalanobis_modified}, we have conducted empirical investigations using different neural network architectures and two different data sets\footnote{Code available on GitHub.}.  The two data sets used are the iris flower for a classification problem and the auto miles per gallon (MPG) for a regression problem. The iris flower data set is a collection of 150 data samples of different iris flowers. Each data sample contains four features: petal length, petal width, sepal length, and sepal width. Based on these four features, the samples are classified into three classes of iris species: setosa, versicolor, and virginica. Figure \ref{fig:iris} plots the data points based on pairwise combinations of the features.

\begin{figure}[!hbtp]
\centering
\includegraphics[width=0.6\textwidth]{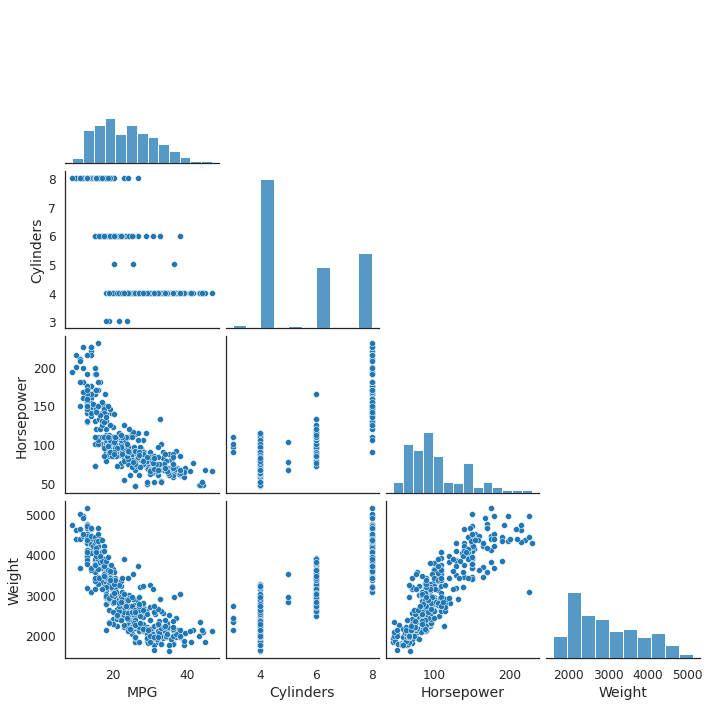}
\caption{Auto MPG data set. Eight attributes (those in bold are shown in the figure): \textbf{number of cylinders}, displacement, \textbf{horsepower}, \textbf{weight}, acceleration, model year, origin, and \textbf{fuel consumption (mpg)}. Off-diagonal graphs are scatterplots of all samples based on pairwise combinations of the attributes shown in the x- and y-axes. For example, looking at MPG vs. Horsepower, one can see that, in general, more power means higher fuel consumption. Diagonal graphs are histograms of the data samples for the attribute shown in the x-axis. For example, one can see that most of the samples have four cylinders, while very few have three and five.}
\label{fig:autompg}
\end{figure}

The auto MPG data set is a collection of 398 data samples of different cars. Each sample contains eight attributes: number of cylinders, displacement (or engine size), horsepower, weight, acceleration, model year, origin, and fuel consumption measured in miles per gallon (mpg). The first seven attributes are then used to predict the fuel consumption. Figure \ref{fig:autompg} plots the data points based on pairwise combinations of four of the attributes.

To maintain focus on the optimization landscape, the neural networks were designed with basic architecture. Layer activations were ReLu, except for the output layer of the regression problem, which had a softmax activation. Dropouts were not used. The optimizer was SGD; however, it should be stressed here that as our focus is on understanding the parameter space, any optimization algorithm could have also been chosen, including gradient descent. As the neural networks were trained, model parameters, i.e., weights and biases, for every iteration $i$ were extracted to create the data sets $X:=\{\mathbf{x}_i\}_{i=1}^M$ containing the points in $\mathbb R^m$ that the optimizer has visited.


\subsection{SGD covariance}
Although SGD was used for optimization, with only 150 samples for the iris flower data set and 398 samples for the auto MPG data set, it was feasible to compute the full gradients in order to calculate the exact SGD covariances. Because of this, we were able to use \eqref{eq:sgd_cov_exact} as the covariances for the Mahalanobis distance in \eqref{eq:mahalanobis_modified} rather than the approximation in \eqref{eq:sgd_cov_approx}, which would have required us to assess where the critical points are and/or would have limited our analysis to data points after convergence.

\begin{figure}[htbp!]
\centering
\begin{subfigure}{.45\textwidth}\label{fig:iris_corr_matrix}
  \centering
  \includegraphics[trim=0cm 0cm 0.5cm 0cm, width=0.9\linewidth]{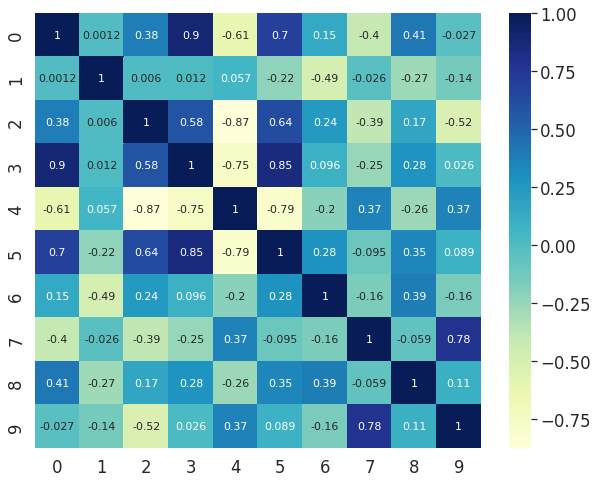}
  \caption{}
\end{subfigure}%
\begin{subfigure}{.45\textwidth}\label{fig:iris_cov_EV}
  \centering
  \includegraphics[trim=0.5cm 0cm 0cm 0cm, width=0.9\linewidth]{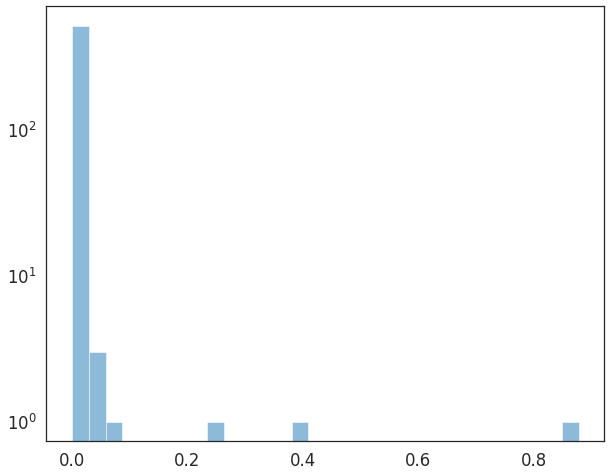}
  \caption{}
\end{subfigure}
\caption{(a) SGD correlation matrix for classification problem (iris data set) at iteration 20 (showing only a 10x10 section of the matrix to zoom in on details). (b) Eigenvalues of the SGD covariance matrix at iteration 20. Note that y-axis is in log scale.}
\label{fig:iris_covariance}
\end{figure}

\begin{figure}[htbp!]
\centering
\begin{subfigure}{.45\textwidth}\label{fig:autompg_cov_matrix}
  \centering
  \includegraphics[trim=0cm 0cm 0.5cm 0cm, width=0.9\linewidth]{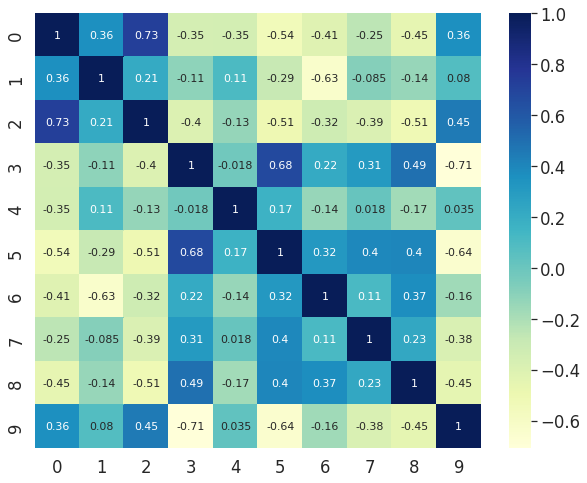}
  \caption{}
\end{subfigure}%
\begin{subfigure}{.45\textwidth}\label{fig:autompg_cov_EV}
  \centering
  \includegraphics[trim=0.5cm 0cm 0cm 0cm, width=0.9\linewidth]{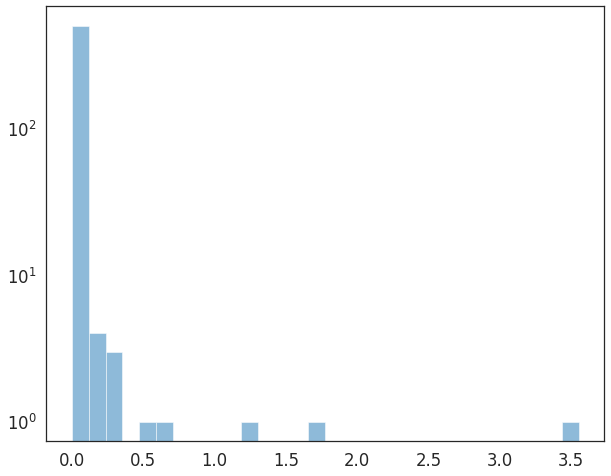}
  \caption{}
\end{subfigure}
\caption{(a) SGD correlation matrix for regression problem (auto MPG data set) at iteration 5 (showing only a 10x10 section of the matrix to zoom in on details). (b) Eigenvalues of the SGD covariance matrix at iteration 5. Note that y-axis is in log scale.}
\label{fig:autompg_covariance}
\end{figure}

Figure \ref{fig:iris_covariance} shows (a) a section of the SGD correlation matrix for the iris flower classification problem at iteration $i = 20$, i.e., at data point $\mathbf{x}_{20}$,  and (b) a histogram of the eigenvalues of the full covariance matrix at the same iteration.
Figure \ref{fig:autompg_covariance}, on the other hand, shows the correlation and histogram of eigenvalues for the auto MPG regression.
10x10 sections were used for the SGD correlation figures instead of the full matrices in order to zoom in on the details and more clearly illustrate the differences in values. These correlation figures, together with the histograms of the covariance eigenvalues, clearly show that the SGD correlations and covariances do not approximate the identity matrix, thus, justifying our choice of using the Mahalanobis distance. If the covariances were close to the identity, this would imply that the Euclidean distance would have been sufficient.

It is also worth noting that, as the covariance is approximately proportional to the Hessian, see \eqref{eq:sgd_cov_hessian}, the values observed for the correlation matrix and covariance eigenvalues, as well as the behavior of the eigenvalues to be concentrated around zero while only a small number of eigenvalues are large, are consistent with the findings of \cite{sagun2017empirical} where the eigenvalues of the Hessian of SGD are examined. These few eigenvalues with large magnitudes represent the principal directions, and are, therefore, the ones of interest.


\subsection{Dimension of subspace}
\label{sec:dim_subspace}
In \cite{Coifman2006}, the number of significant eigenvalues $s$ is defined through a preset accuracy $\alpha > 0$ on which $s$ depends,
\begin{equation} \label{eq:coifman_dim}
    s(\alpha) := \max\{l\in \N: {|\lambda_l|} > \alpha{|\lambda_1|}\}.
\end{equation}
$s=s(\alpha)$ is then interpreted as the dimension of the underlying slow manifold. The original data set is embedded into the new $s(\alpha)$-dimensional subspace through the mapping in \eqref{eq:diffusion_map}, where the diffusion distance is equal to the Euclidean distance up to the relative accuracy $\alpha$. That is, with $\tau = 1$ in \eqref{eq:diffusion_distance1},
\begin{equation}
\label{eq:diffusion_distance_approx}
d(\mathbf{x}_i,\mathbf{x}_j) = \Bigg(\sum_{l=1}^{s(\alpha)} \lambda^{2}_\ell(\psi_\ell(\mathbf{x}_i)-\psi_\ell(\mathbf{x}_j))^2\Bigg)^{1/2}.
\end{equation}
$\alpha$ is a parameter that needs to be selected, where a smaller $\alpha$ leads to higher dimensions of the slow manifold and higher accuracy in \eqref{eq:diffusion_distance_approx}.

There is, however, not one unique way to define the intrinsic dimension. In our approach, we look at what we refer to as the energy ratio, $\Lambda_i$, defined as
\begin{align}
    \Lambda_i:=\frac{\sqrt{\lambda_1^2+\lambda_2^2+\dots+\lambda_i^2}}{\sqrt{\lambda_1^2+\lambda_2^2+\dots+\lambda_N^2}}.
\end{align}
This ratio quantifies the dominance of the first $i$ eigenvalues by comparing their energy with the total energy of all eigenvalues. Figures \ref{fig:ev_iris&auto} and \ref{fig:ev_inits} in Section \ref{sec:results} show graphs of $\Lambda_i$ for different neural networks models. We define the number of significant eigenvalues, denoted by $d$, i.e., the dimension of the underlying lower dimensional subspace, to be the first $d$ eigenvalues such that $\Lambda_d>\beta$. A lower $\beta$ results in lower dimensions, and we, in the interest of employing a strict criteria, choose $\beta = 0.99$. Looking at $\Lambda_i$ and defining the significant eigenvalues as such captures the spectrum decay of the matrix $P$ (see Section \ref{sec:diffusion_kernelmap}), as does \eqref{eq:coifman_dim}, upon which the dimension of the subspace depends. In addition, we also examine the proportion of the area under the eigenvalue curve accounted for by the dominant $d$ eigenvalues. We refer to this as the AUC ratio and it can be interpreted as the explanatory capability of, or the amount of information contained in, the lower dimensional subspace in comparison to the original space.


\subsection{Diffusion map parameter $\varepsilon$}\label{sec:diff_map_epsi}
A significant parameter in the implementation of diffusion maps is the scale parameter $\ve$ used in the definition of the diffusion kernel in \eqref{k_epsilon}.  It represents a characteristic distance in the data and defines the local neighborhood within which we can rely on the accuracy of our metric (Mahalanobis distance in this case). Results can vary tremendously depending on its setting. Despite the importance of the parameter and the sensitivity of results, there is no agreed upon scheme as to how the appropriate range of values should be decided. Instead, the choice is dependent upon the problem and the data structure, resulting in different methods being proposed. For example, in \cite{LafonThesis}, $\ve$ is set to be
\begin{equation*}
\label{eq:lafon_epsilon}
    \varepsilon = \frac{1}{M}{\sum_{i=1}^M \min_{j:j \neq i}\Vert \mathbf{x}(t_i)-\mathbf{x}(t_j)\Vert^2},
\end{equation*}
which is the average of the shortest distance from each data point. Implementing this on our data using the Mahalanobis distance, however, resulted in an $\varepsilon$ that was too small compared to other values of $\Vert \mathbf{x}(t_2)-\mathbf{x}(t_1)\Vert_{MD}^2$. There were very few, if any, data points within the ball of radius $\ve$, and many entries of the $K$ matrix were almost zero. In \cite{Coifman2016}, the authors looked at the error $E_{MD}(\mathbf{y}(t_1),\mathbf{y}(t_2))$ incurred by using the Mahalanobis distance on the data points $\mathbf{y}(t) = f(\mathbf{x}(t))$ in approximating the $L_2$-distance of the underlying variables. $\mathbf{x}(t)$ at times $t_1,\ldots, t_n$ are the samples of the stochastic system. The criteria used is that $\varepsilon$ should be in the order of $\Vert \mathbf{x}(t_2)-\mathbf{x}(t_1)\Vert_{MD}^2$ in the region where $|E_{MD}(\mathbf{x}(t_1),\mathbf{x}(t_2)| \ll \Vert \mathbf{x}(t_2)-\mathbf{x}(t_1)\Vert_{MD}^2$. Choosing $\varepsilon$ as such ensures that the curvatures and nonlinearities captured in the error term are negligible. This method, however, is inapplicable for our investigation as we are interested in the stochastic variable itself, not in the underlying variable, and thus we have no error term to consider. On the other hand, in \cite{bah2008diffusion, coifman2008graph, singer2009detecting}, the authors calculate the matrix  $K(\varepsilon)$ for a wide range of $\varepsilon$ values and compute the sum $L(\ve)$ of the entries for each matrix:
\begin{equation*}
\label{eq:singer_epsilon}
    L(\varepsilon) = \sum_{i,j} K_\varepsilon[i,j].
\end{equation*}
An $\varepsilon$ that is too small compared to $\Vert \mathbf{x}(t_2)-\mathbf{x}(t_1)\Vert_{MD}^2$ will result in a lower value for $L(\varepsilon)$, since the entries for the matrix $K$ will be close to zero, indicating little to no diffusion. In contrast, an $\varepsilon$ that is  too large compared to $\Vert\mathbf{x}(t_2)-\mathbf{x}(t_1)\Vert_{MD}^2$ will result in a larger $L(\varepsilon)$, as the entries of $K$ will be close to one, indicating that diffusion has already taken place. Since neither of these scenarios are interesting for diffusion maps, $\varepsilon$ should be chosen in the region between. In \cite{coifman2008graph}, assuming that the data points lie on a low-dimensional manifold $\mathcal{M}$ with finite volume, it is argued that the sum  $L(\varepsilon)$ is approximated by the mean value integral. That is,
\begin{equation}
     L(\varepsilon) = \sum_{i,j} K_\varepsilon[i,j] = \sum_{i,j} \exp\left(- \frac{\Vert\mathbf{x}_i-\mathbf{x}_j\Vert^2}{\varepsilon}\right) \approx \frac{N^2}{\textrm{vol}^2(\mathcal{M})} \int_\mathcal{M} \int_\mathcal{M} \exp\left(- \frac{\Vert\mathbf{x}-\mathbf{y}\Vert^2}{\varepsilon}\right)d\mathbf{x}d\mathbf{y}.
\end{equation}
Since the manifold $\mathcal{M}$ looks like its tangent space $\mathbb{R}^d$ locally,

\begin{align}
\begin{split}
    \frac{N^2}{\textrm{vol}^2(\mathcal{M})} \int_\mathcal{M} \int_\mathcal{M} \exp\left(- \frac{\Vert\mathbf{x}-\mathbf{y}\Vert^2}{\varepsilon}\right)d\mathbf{x}d\mathbf{y}
    &\approx \frac{N^2}{\textrm{vol}^2(\mathcal{M})} \int_\mathcal{M} \int_{\mathbb{R}^d} \exp\left(- \frac{\Vert\mathbf{x}-\mathbf{y}\Vert^2}{\varepsilon}\right)d\mathbf{x}d\mathbf{y} \\
    &= \frac{N^2}{\textrm{\textrm{vol}}(\mathcal{M})}(\pi\varepsilon)^{d/2}.
\end{split}
\end{align}
Taking the logarithm,
\begin{equation}
\label{eq:linear_epsilon}
    \textrm{log}L(\varepsilon) \approx \frac{d}{2}\textrm{log}\varepsilon + \textrm{log}\left(\frac{N^2\pi^{d/2}}{\textrm{vol}(\mathcal{M})}\right).
\end{equation}
Here, $\textrm{vol}(\mathcal{M})$ is the volume of the manifold. The logs of $L(\ve)$ and $\ve$ are, therefore, connected by an approximately straight line whose slope is $d/2$, where $d$ is the dimension of the lower-dimensional manifold. The authors suggest to choose an $\varepsilon$ within this linear region. \cite{berry2016variable} extends this approach further by setting $\ve$ to be where the local slope $a_i$, given approximately by
\begin{equation}
\label{eq:dim_slope}
    a_i \approx \frac{\textrm{log}(L(\varepsilon_{i+1})) - \textrm{log}(L(\varepsilon_i))}{\textrm{log}(\varepsilon_{i+1}) - \textrm{log}(\varepsilon_i)},
\end{equation}
is maximized. In this case, the slope $d/2 \approx \textrm{max}\{a_i\}$, and, hence, the dimension of the manifold is given by $d \approx 2\textrm{max}\{a_i\}$.

\begin{figure}[htbp!]
\centering
\includegraphics[width=1\textwidth]{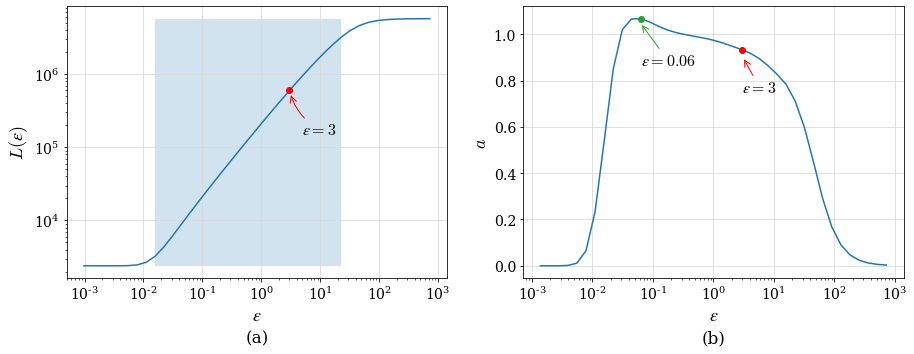}
\caption{(a) As per $\ve$-choosing criteria of \cite{coifman2008graph, singer2009detecting}. Log-log plot of $\varepsilon$ against $L(\varepsilon)$ showing linear region (shaded in blue) within which $\varepsilon$ should be chosen. (b) As per $\ve$-choosing criteria of \cite{berry2016variable}. Graph of $\varepsilon$ and slope $a$. Green dot shows where the slope is maximized, corresponding to $a = 1.07$. Red dots in both graphs show the actual chosen $\varepsilon = 3$. From (a), it shows that $\varepsilon = 3$ is within accepted region, and from (b), although not the maximum, $\varepsilon = 3$ still gives a high value of $a$.}
\label{fig:epsi_line&a}
\end{figure}

Figure \ref{fig:epsi_line&a} shows the results of implementing the criteria of \cite{coifman2008graph, singer2009detecting, berry2016variable} for our data set $X:=\{\mathbf{x}_i\}_{i=1}^M$ of points in $\mathbb R^m$ visited by SGD. The iris flower classification problem with a two-hidden layer neural network was used, with an original parameter space dimension of $m = 515$. Section \ref{sec:class_vs_reg} describes this neural network in more detail.
Figure \ref{fig:epsi_line&a}(a) is the log-log plot of $\varepsilon$ against $L(\varepsilon)$, showing where $L(\varepsilon)$ grows linearly with $\varepsilon$. According to \cite{coifman2008graph} and \cite{singer2009detecting}, this is the optimal region within which $\varepsilon$ should be chosen. Figure \ref{fig:epsi_line&a}(b) shows $\ve$ with corresponding slopes $a$. The green dot is where the maximum is attained, with $a = 1.07$ and $\ve = 0.06$. Based on \cite{berry2016variable}'s definition, the dimension of the lower dimensional subspace of the parameter space is then $d = 2$. To err on the conservative side, this result appeared to be very optimistic. In addition, if we, instead, use our definition of dimension in Section \ref{sec:dim_subspace}, using $\varepsilon = 0.06$ in the diffusion maps resulted in $d = 868$, meaning that not even the dimension of the original parameter space was recovered.

\begin{figure}[htbp!]
\centering
\includegraphics[width=1\textwidth]{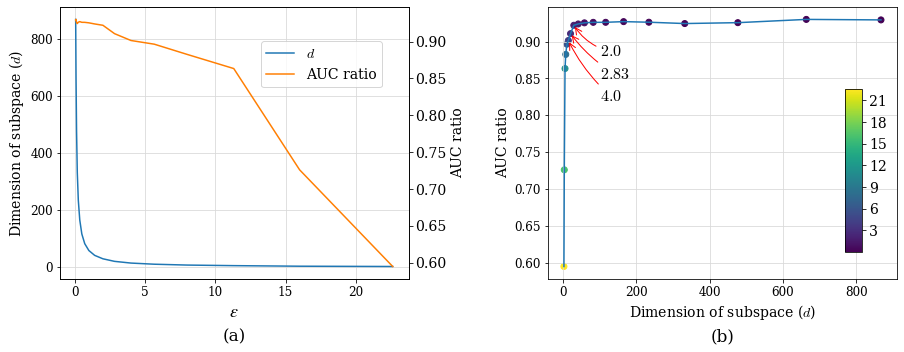}
\caption{(a) Dimension of slow manifold (in blue and left y-axis) and AUC ratio (in orange and right y-axis) as functions of $\varepsilon$. (b) Scatterplot of different values of $\varepsilon$, showing resulting dimension in x-axis and AUC ratio in y-axis from application of diffusion maps. The color bar shows the range of values for the $\varepsilon$'s. $\varepsilon$'s in the upper left corner are desirable due to low-dimension and high AUC ratio. Several of these points are marked simply for reference.}
\label{fig:epsi_iris_zoomed}
\end{figure}

As none of these methods showed reasonable results for our problem, we instead modified. Instead of choosing $\varepsilon$ where the slope is maximized, we instead studied the range of possible values of $\varepsilon$ in the linear region showed in Figure \ref{fig:epsi_line&a}(a). Figure \ref{fig:epsi_iris_zoomed} shows the AUC ratios and dimensions, as defined in Section \ref{sec:dim_subspace}, for these values.
The results presented in this section are those of the iris flower classification problem, although the same approach was also applied for the auto MPG regression. As $\ve$ increases, the dimension of the underlying subspace decreases sharply in the beginning, then "flattens out" to converge to 1, as shown in Figure \ref{fig:epsi_iris_zoomed}(a). The AUC ratio also decreases as $\ve$ increases, which is expected as the dimensions decrease. However, the decrease in AUC ratio is not nearly as abrupt as that of the dimensions. The decrease is subtle for smaller values of $\ve$ and gradually becomes steeper as $\ve$ increases. The sudden decline in dimensions coupled with just a slight decrease in AUC ratio indicates that the decline in dimension is the result of better detection of the lower dimensional subspace due to better parameterization of the data, and that these detected lower dimensional subspaces do, in fact, account for a lot of the information in the original optimization landscape. We then choose $\varepsilon$ based on Figure \ref{fig:epsi_iris_zoomed}(b), where values on the upper left corner are desired as they result in a combination of lower dimensions and higher AUC ratios. After careful assessment, the value of the parameter was decided to be $\varepsilon = 3$ for the classification problem and $\varepsilon = 0.55$ for the regression problem. Figure \ref{fig:epsi_line&a} shows $\varepsilon = 3$ marked as a red dot in order to show that this choice does indeed fall within the linear region and, although not the maximum, does still correspond to a high slope $a$. Note that as there is not one optimal value but rather a range of accepted values, other surrounding values of the same order would have also been suitable. With these values of $\varepsilon$, the diffusion maps were applied to the data set $X$. The eigenvalues were calculated as described in Section \ref{sec:diffusion_kernelmap}, and sorted in descending order $\lambda_1\geq \lambda_2\ge \dots$.


\subsection{Results} \label{sec:results}
In our empirical investigations, we tried to ascertain whether or not the high-dimensional parameter surface does indeed have an underlying low-dimensional manifold in which the process of minima selection occurs, and, if so, determine what variables affect the dimensions of this manifold. Due to the randomness of SGD, the experiments were repeated multiple times, and the resulting dimensions as well as AUC ratios have proven to be stable throughout. For example, 30 different runs of the classification problem described in Section \ref{sec:class_vs_reg} consistently detected a subspace of dimension $d = 17$ with  $AUC \approx 0.91$. The results are presented in detail in the following sections.

\subsubsection{Classification and regression} \label{sec:class_vs_reg}
To first possibly detect the lower dimensional subspace, standard neural networks were applied to the iris flower and auto MPG data sets. Using these data sets allowed us to look at both classification and regression problems, and thus, different loss functions. Categorical cross entropy (CCE) was the loss function for the classification problem
\begin{equation}
\label{cross_ent}
f(\mathbf{x})=-\frac1{N} \sum\nolimits_{i=1}^N\sum\nolimits_{j=1}^c y_{ij}\log(\hat{y}_{ij}),
\end{equation}
where $c$ is the number of classes, and $y_{ij}$ and $\hat{y}_{ij}$ are the true and predicted labels for sample $i$ and class $j$, respectively. Mean absolute error (MAE) was the loss function for the regression,
\begin{equation}
\label{mae}
f(\mathbf{x}) =  \frac1{N} \sum\nolimits_{i=1}^N |\y_{i}-\hat{y}_{i}|,
\end{equation}
where $y_{i}$ are the true labels, and $\hat{y}_i$ are the predicted labels. For comparability, the architectures were designed to be as similar as possible. Both had two hidden layers with neurons 24 and 14 for the iris flower, and 21 and 13 for the auto MPG. These were chosen so that the neural networks would have similar width and depth, as well as the same parameter space dimension of $m = 515$. In addition, both were trained for the same batch size of $n = 20$, with 400 epochs for the iris and 150 for the auto MPG so as to have the same number $M$ of SGD steps. Table \ref{table:ev_iris&auto} summarizes these setup together with the results.

\begin{figure}[htbp!]
\centering
\includegraphics[width=1\textwidth]{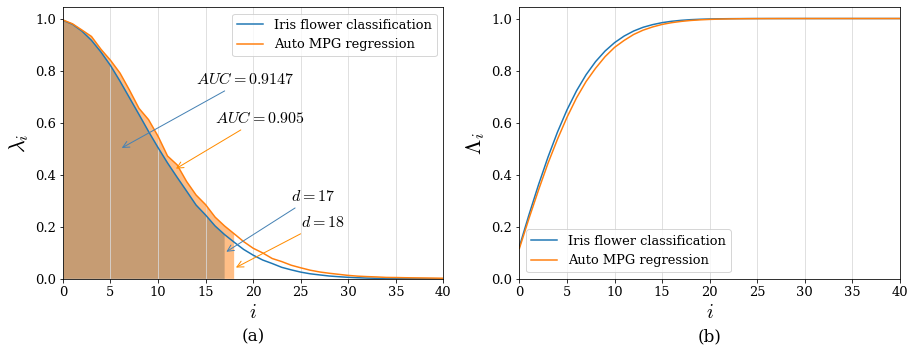}
\caption{(a) Eigenvalues $\lambda_i$ for the iris flower classification and auto MPG regression problems. AUC ratios and dimensions of lower subspaces annotated. (b) Energy ratios $\Lambda_i$ for the classification and regression problems. For both, as $i$ increases, $\lambda_i \to 0$ and $\Lambda_i \to 1$.}
\label{fig:ev_iris&auto}
\end{figure}

Figure \ref{fig:ev_iris&auto} plots the eigenvalues $\lambda_i$  and energy ratios $\Lambda_i$ resulting from the application of diffusion maps. The dimensions $d$ of the subspaces as well as the AUC ratios are annotated in the graphs.
For both models, only a small number of eigenvalues are actually dominant, while the others may be considered insignificant. As the eigenvalues indicate the importance of their associated direction, this result does indeed support the hypothesis that the SGD optimizer moves in a lower-dimensional subspace. For the iris flower classification, the dimension of this subspace appears to be $d=17$, while for the auto MPG regression, it is $d=18$. The classification problem has a slightly higher AUC ratio of 0.9147, with the regression having 0.9050. Both of these ratios indicate high amounts of information in these lower dimensional subspaces.
\begin{table}[htbp!]
\centering
\begin{tabular}{c|cccccccc}
\hline
Model       &   Hidden Layers  & Neurons & $N$&    $n$   & $M$ &$m$       & $d$    & AUC Ratio
\\\hline
Iris flower classification &   2 & 24x14& 150&20&	2400 & 515  & 17  & 0.9147		\\
Auto MPG regression         &   2 & 21x13& 398&20&   2400 & 515  & 18     & 0.9050    \\
\hline
\end{tabular}
\caption{Summary of the number of samples $N$, batch size $n$, number of SGD data points $M$, dimension of original parameter space $m$, dimension of subspace $d$ detected by diffusion mapping, and AUC ratios.}
\label{table:ev_iris&auto}
\end{table}

\begin{figure}[hbtp!]
\centering
\includegraphics[width=0.5\textwidth]{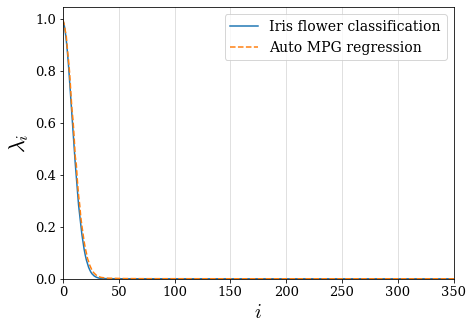}
\caption{Eigenvalues resulting from diffusion maps graphed until $i = 350$ to show that they exhibit sharp decay.}
\label{fig:ev_iris&auto_decay}
\end{figure}

It is also worth noting that the graphs in Figure \ref{fig:ev_iris&auto} are cut off at $i = 40$ in order to zoom in on the details. Beyond $i = 40$, $\lambda_i \to 0$ and $\Lambda_i \to 1$. This means that, when graphed together with all the eigenvalues, a very sharp decay in the spectrum can be observed. Figure \ref{fig:ev_iris&auto_decay} graphs the eigenvalues up until $i = 350$ to illustrate this point.

\subsubsection{Batch size}
With the lower dimensional manifold being detected in Section \ref{sec:class_vs_reg}, it is interesting to examine what variables influence its dimensions. The batch size is known to be an important hyperparameter to tune for neural network models. To analyze its effect on the dimension, the iris flower classification problem described in Section \ref{sec:class_vs_reg} was used. The batch size was varied from 10 to 120, where 120 corresponds to a full gradient descent\footnote{The iris flower data set has 150 samples, but only 120 was used for training as the rest was used for validation.}. Batch sizes by which the number of training data is divisible were intentionally chosen to ensure no batches with remainders are left at the end of epochs. Results, displayed in Figure \ref{fig:batch_depth}(a) and Table \ref{table:batch}, show that as the batch size is varied, the dimension of the subspace and AUC ratios remain the same. Also, the dimension of the noisy SGD is equal to the dimension of the full gradient descent, indicating the dimension's robustness to noise. This is because even though decreasing the batch size increases the noise in the SGD, the noise still have the same directions, just different magnitudes. Hence, the slow variables in the parameter surface, and, thus, the dimension of the subspace, are unchanged.
\begin{table}[hbt!]
\centering
\begin{tabular}{c|cc}
\hline
Batch size ($n$)        & $d$   & AUC Ratio     \\\hline
10	&	17	&	0.9140\\	
20	&	17		&	0.9147	\\
30	&	17		&	0.9145	\\
40	&	17		&	0.9147	\\
60	&	17		&	0.9147	\\
120 (full batch)	&	17		&	0.9147	\\
\hline
\end{tabular}
\caption{Subspace dimensions $d$ and AUC ratios for varying batch size $n$. Results show that dimension and AUC ratio remain consistent regardless of batch size.}
\label{table:batch}
\end{table}

\subsubsection{Neural network depth}
\begin{figure}[hbt!]
\centering
\includegraphics[width=1\textwidth]{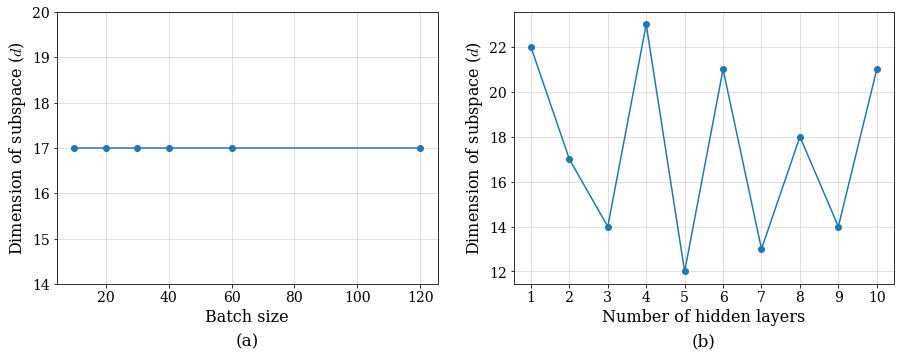}
\caption{(a) Subspace dimension vs. batch size. Dimension remains consistent regardless of batch size. (b) Subspace dimension vs. number of hidden layers. Dimension fluctuates around $d = 17$.}
\label{fig:batch_depth}
\end{figure}

The dimension of the subspace as a function of depth was also examined. Once again, the iris flower classification problem from was used. Layers and neurons were adjusted to increase depth, while keeping the number of model parameters $m$ similar. The results are graphed in Figure \ref{fig:batch_depth}(b) and summarized in Table \ref{table:depth}.
\begin{table}[htb!]
\centering
\begin{tabular}{c|p{0.26\linewidth}cccc}
\hline
Hidden Layer      & Neurons & $m$   & $d$   & AUC Ratio & Loss   \\\hline
1	&	64	&	515	&		22	&	0.9145  &0.0987\\	
2	&	24, 14	&	515	&		17		&	0.9147	&0.0686\\
3	&	12, 18, 10	&	517	&		14		&	0.9044	&0.0594\\
4	&	10, 14, 12, 8	&	515	&		23		&	0.9146	&0.0641\\
5	&	9, 10, 13, 10, 6	&	515	&		12		&	0.9067	&0.0514\\
6	&	6, 8, 12, 12, 8, 5	&	517	&		21		&	0.9120	&0.0502\\
7	&	5, 8, 9, 11, 9, 8, 5	&	515	&		13		&	0.9019	&0.0513\\
8	&	5, 7, 9, 10, 9, 8, 6, 4	&	515	&		18		&	0.9108	&0.1060\\
9	&	5, 6, 7, 8, 10, 8, 7, 6, 5	&	516	&		14		&	0.9006	&0.0434\\
10	&	4, 5, 6, 8, 9, 9, 8, 6, 5, 4	&	516	&		21		&	0.9151	&0.4665\\
\hline
\end{tabular}
\caption{Dimension of subspace $d$ and AUC ratios for different number of hidden layers. Loss after 400 epochs also displayed.}
\label{table:depth}
\end{table}
One can see that the depth of the neural network does have an influence on the dimension. However, it is unclear as to what the relationship is between the two. The dimensions seem to fluctuate around $d = 17$ as the hidden layer is increased. This may indicate that the intrinsic dimensionality of the manifold in which SGD moves around depends more on the data set rather than the neural network architecture. It also appears as though the suitability of the number of hidden layers does not affect the dimension either. The loss column in \ref{table:depth} was included to illustrate this point. The neural network with ten hidden layers, for example, has a much higher loss than all other models in the table after being trained for the same number of epochs. This higher loss signifies that this neural network model is ill-designed for the problem. However, the detected dimension is still $d = 21$, the same as the six-hidden neural network that performed better in training. Again, this supports the conjecture that the dimensionality relies more on the data set, and is therefore insensitive, to some degree, to depth or suitability of network architecture.
Another interesting observation is that, apart from the one-hidden layer model, all other odd-hidden-layered models appear to have lower dimensions that those that are even. A more thorough investigation, however, needs to be conducted in order to make any conclusions.

\subsubsection{Weight Initializations}
\begin{figure}[hbp!]
\centering
\includegraphics[trim=0cm 0cm 0cm 0cm, width=\textwidth]{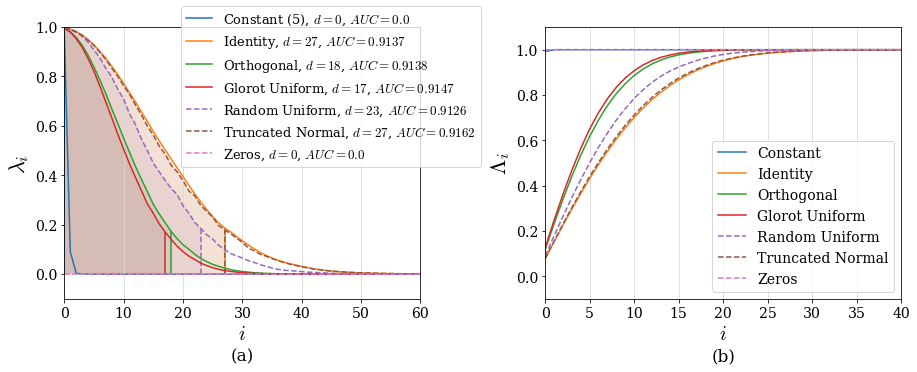}
\caption{Eigenvalues $\lambda_i$ and energy ratios $\Lambda_i$ for different weight initializations. Note that in (a), the eigenvalues for constant and zeros are very close to $\lambda_i = 0$, and in (b), the energy ratios for constant and zeros are very close to $\Lambda_i = 1$.}
\label{fig:ev_inits}
\end{figure}

\begin{figure}[hbt!]
\centering
\includegraphics[width=\textwidth,center]{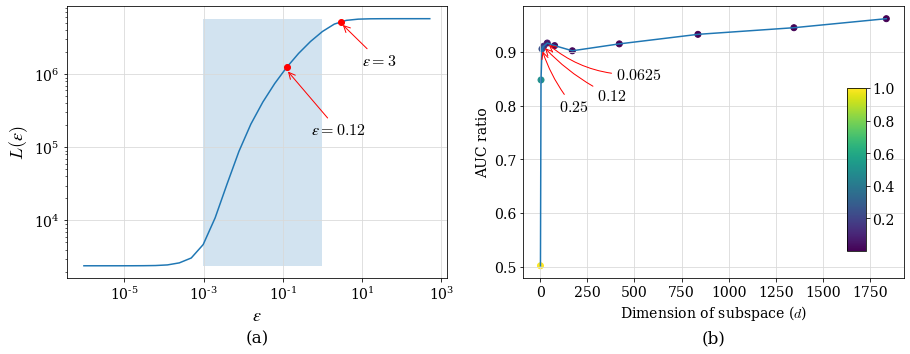}
\caption{Re-tuning of scale parameter $\varepsilon$ for $\textrm{constant} ( = 5)$ initialized model. (a) Optimal region shaded in blue within which $\ve$ should be chosen. Chosen value of $\ve = 0.12$ and previously chosen value of $\ve = 3$ marked as red dots for comparison. (b) Possible optimal values based on graph of dimension $d$ against AUC ratio. The colorbar shows the range of values for $\ve$. Values in the upper left corner are desirable due to low-dimension and high AUC ratio. Several points marked simply for reference.}
\label{fig:epsi_const}
\end{figure}

\begin{figure}[hbt!]
\centering
\includegraphics[width=\textwidth,center]{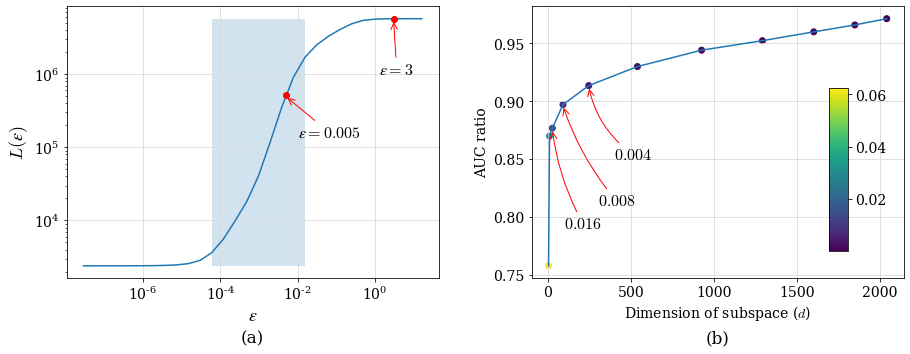}
\caption{Re-tuning of scale parameter $\varepsilon$ for zeros initialized model. (a) Optimal region shaded in blue within which $\ve$ should be chosen. Chosen value of $\ve = 0.005$ and previously chosen value of $\ve = 3$ marked as red dots for comparison. (b) Possible optimal values based on graph of dimension $d$ against AUC ratio. The colorbar shows the range of values for $\ve$. Values in the upper left corner are desirable due to low-dimension and high AUC ratio. Several points marked simply for reference.}
\label{fig:epsi_zeros}
\end{figure}

In the preceding sections, the weight initializations were kept to be Keras' default Glorot Uniform\footnote{Uniform Distribution$[-limit,limit]$, where $limit = \sqrt{\frac{6}{fan\_in + fan\_out}}$. $fan\_in$ and $fan\_out$ are the numbers of input and output units to the weight tensor, respectively.} for consistency. However, weight initialization is also an interesting variable to study when determining possible factors of dimension. As before, the iris flower with two hidden layers described in Section \ref{sec:class_vs_reg} was used, and different weight initializations available in Keras were implemented. Figure \ref{fig:ev_inits} plots the eigenvalues and energy ratios.
Notice that, for the constant\footnote{The constant initialization was set to be = 5.} and zero initializations, $d = 0$ and $AUC = 0$. These results are due to the value of the scale parameter, $\varepsilon = 3$, not being suitable for those initializations. Consequently, $\ve$ had to be adjusted using the method detailed in Section \ref{sec:diff_map_epsi}. Figures \ref{fig:epsi_const} and \ref{fig:epsi_zeros} plot (a) the approximately linear regions within which $\ve$ should be chosen for these two initializations, and (b) possible optimal values based on the graphs of dimension $d$ against the AUC ratio.
$\ve = 3$ is marked as a red dot in the graphs to show that this value is clearly unfitting. More appropriate choices of $\varepsilon = 0.12$ for constant and $\varepsilon = 0.005$ for zeros were made. With these new values, the lower dimensional manifolds were more correctly uncovered with $d=20$ for the constant initialization and $d=178$ for the zero initialization. Table \ref{table:auc_inits} summarizes the results.
\begin{table}[hbp!]
\centering
\begin{tabular}{l|cccc}
\hline
Initialization      & $m$   & $M$   & $d$ & AUC Ratio     \\\hline
Constant (=5, $\varepsilon = 3$)	    &	515	&	2400	&	0	&	0.0000	\\
Constant (=5, $\varepsilon = 0.12$)	    &	515	&	2400	&	20	&	0.9150	\\
Identity	                            &	515	&	2400	&	27	&	0.9137	\\
Orthogonal	                            &	515	&	2400	&	18	&	0.9138	\\
Glorot Uniform	                        &	515	&	2400	&	17	&	0.9147	\\
Random Uniform	                        &	515	&	2400	&	23	&	0.9126	\\
Truncated Normal	                    &	515	&	2400	&	27	&	0.9162	\\
Zeros ($\varepsilon = 3$)	            &	515	&	2400	&	0	&	0.0000	\\
Zeros ($\varepsilon = 0.0005$)	        &	515	&	2400	&	178	&	0.9098	\\
\hline
\end{tabular}
\caption{Summary of subspace dimensions $d$ and AUC ratio for different weight initializations.}
\label{table:auc_inits}
\end{table}
Significant reductions in dimension for all initializations can be observed. Only for very poorly initialized models, such as the case with zeros where $d = 178$, does SGD fail to find a much lower dimensional subspace in which to move around. This signifies that, although initial weights still should be selected carefully to ensure that SGD performs well, the dimension of the subspace does appear to be robust to initialization.

\subsubsection{Convergence and stability}
\begin{figure}[htbp!]
\centering
\begin{subfigure}{.46\textwidth}\label{fig:conv_L6_loss}
  \centering
  \includegraphics[trim=2cm 0.5cm 0cm 0.5cm,width=0.83\textwidth,center]{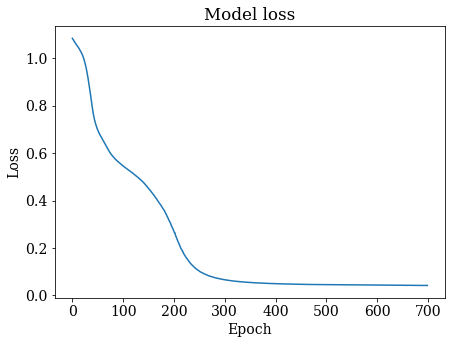}
  \caption{}
\end{subfigure}%
\begin{subfigure}{.46\textwidth}\label{fig:conv_iris_fixed}
  \centering
  \includegraphics[trim=0cm 0.5cm 0cm 0cm, width=1.1\textwidth,center]{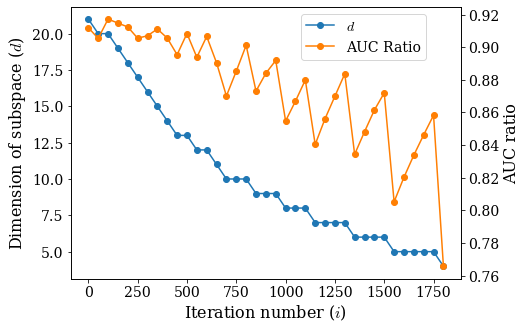}
  \caption{}
\end{subfigure}
\caption{(a) Loss graph for the iris flower classification problem with six hidden layers (see details in Table \ref{table:depth}). SGD finds a potential minimum around $\textrm{loss} \approx 0.6$, then escapes to find a better minimum closer to zero before converging. (b) SGD seemingly finding lower dimensional subspaces as optimization continues. Result is erroneous, however, as it results from using a fixed $\ve$ that becomes too large for later data points.}
\label{fig:conv_loss&fixed}
\end{figure}

\begin{figure}[htbp!]
\centering
\includegraphics[width=0.9\textwidth,center]{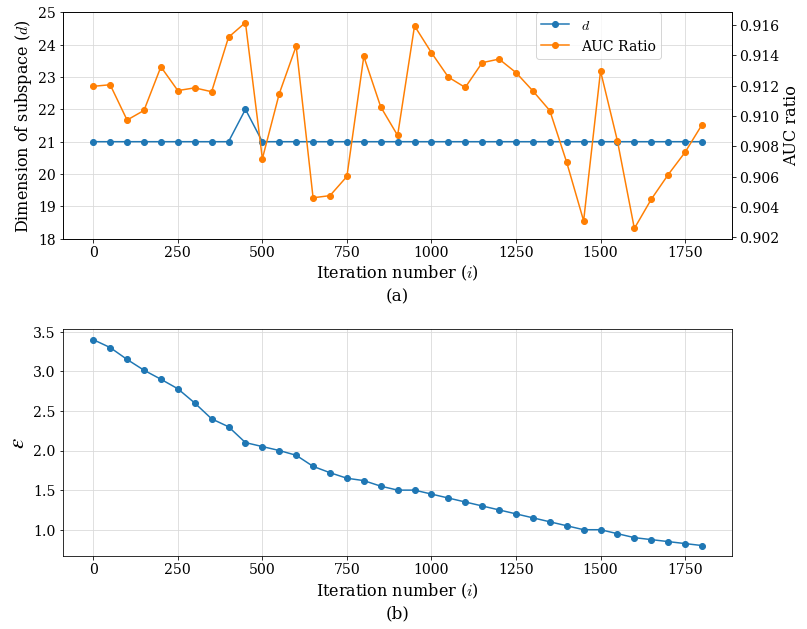}
\caption{(a) Subspace dimension  $d$ (in blue and left y-axis) and AUC ratio (in orange and right y-axis) as SGD progresses. (b) Readjustments made for the scale parameter $\varepsilon$. Keeping parameter fixed from the start means the value becomes too large for later SGD steps.}
\label{fig:conv_iris}
\end{figure}

Apart from attempting to identify the variables that affect the subspace dimension, we also briefly examined how fast SGD moves into the subspace and whether it proceeds to find even lower dimensional subspaces as it continues. Rather than the two-hidden layer model for the iris flower classification that has been used previously, we looked at the model with six hidden layers (see details in Table \ref{table:depth}). The reason for this is that the optimization for this model has more interesting developments, shown in Figure \ref{fig:conv_loss&fixed}(a), where SGD first finds a potential minimum around $\textrm{loss} \approx 0.6$, then escapes to find a better minimum closer to zero before converging.
The first 2400 points were studied to determine the dimension at the beginning of optimization, then a window of the same number of points, but incremented by 50 is rolled to evaluate how the dimension changes. Figure \ref{fig:conv_iris}(a) displays the results, showing that SGD moves into the lower dimensional subspace after just a few steps, and is stable as it stays in the same dimensional subspace even as it finds minima and converges. Figure \ref{fig:conv_iris}(b) also shows how $\varepsilon$ was readjusted.
The scale parameter $\ve$ had to be re-tuned accordingly as SGD converges and the steps become closer together. Maintaining the same $\ve$ throughout means that the parameter becomes too large for later data points, allowing too many adjacent points and noise within the ball of radius $\ve$. This mislead to results that SGD converges to lower and lower dimensions as shown in Figure \ref{fig:conv_loss&fixed}(b).

\section{Summary, conclusions and future research}\label{sumu}
In this paper  we  pursued a truly data driven approach to the problem of getting a potentially deeper understanding of the high-dimensional parameter loss surface, and the landscape traced out by SGD, in the context of fitting (deep) neural networks to real data sets and by analyzing the data generated through SGD in order to possibly discovery (local) low-dimensional representations of the optimization landscape. As our vehicle for the exploration
we used diffusion maps introduced by R. Coifman and coauthors. Our empirical results suggest that the high-dimensional loss surface does indeed contain a lower dimensional subspace in which SGD tends to concentrate/move. The dimension of this subspace is quite significantly lower compared to the dimension of  the loss surface. For example, in the case of the two-hidden layer iris flower model studied, the original parameter space has a dimension of $515$, while the subspace defined has a dimension of $d = 17$, corresponding to an approximately 97\% reduction of dimensionality.  In fact, our results may lean to the conservative side,  as other approaches to defining the subspace, its dimension and the criteria for choosing $\ve$, lead to even lower dimensions, see Sections \ref{sec:dim_subspace} and \ref{sec:diff_map_epsi}. Our empirical results also indicate that the dimension of the subspace defined is, to some degree, robust to the noise, depth, and weight initialization of the neural network. This possibly indicates that the intrinsic dimensionality may be more dependent on the data set rather than the neural network architecture. Moreover, SGD exhibits stability by moving to the lower dimensional subspace just after a few steps, and remains in the subspace as optimization continues.

We think that our empirical results could constitute the beginning of  more comprehensive studies of this interesting and relevant research problem. Finding the relationship between dimension of the subspace introduced and potential factors, in the data or in the architectures used, is complex, and further endeavors should look into different variables in order to attempt to make stronger conclusions as to what really influences the dimension of the subspace. The width of the neural network, for example, can be examined, as well as the use of more data points and different types of larger data sets, the size of which in this study was restricted  due to computational resource constraints. To take full advantage of diffusion maps, subsequent investigations should also examine the actual embedding and determine characterizations of the lower dimensional subspace. Furthermore, in this paper we have focused on SGD for high dimensional problems in the context of deep learning. However, it would also be worthwhile to apply the same methods and examine the behavior of SGD in other high dimensional settings such as in latent factor models where SGD is widely adapted as a learning algorithm.

\bibliographystyle{abbrv}
\bibliography{Bibliography}

\begin{thebibliography}{10}

\bibitem{bah2008diffusion}
B.~Bah.
\newblock Diffusion maps: analysis and applications.
\newblock Master's thesis, Oxford University, 2008.

\bibitem{berry2016variable}
T.~Berry and J.~Harlim.
\newblock Variable bandwidth diffusion kernels.
\newblock {\em Applied and Computational Harmonic Analysis}, 40(1):68--96,
  2016.

\bibitem{chaudhari2017entropy}
P.~Chaudhari, A.~Choromanska, S.~Soatto, Y.~LeCun, C.~Baldassi, C.~Borgs,
  J.~Chayes, L.~Sagun, and R.~Zecchina.
\newblock Entropy-sgd: Biasing gradient descent into wide valleys.
\newblock {\em Journal of Statistical Mechanics: Theory and Experiment},
  2019(12):124018, 2019.

\bibitem{choromanska2015loss}
A.~Choromanska, M.~Henaff, M.~Mathieu, G.~Ben~Arous, and Y.~LeCun.
\newblock {The Loss Surfaces of Multilayer Networks}.
\newblock {\em Proceedings of the 18th International Conference on Artificial
  Intelligence and Statistics}, 38:192--204, May 2015.

\bibitem{Coifman:2014}
R.~R. Coifman and M.~J. Hirn.
\newblock Diffusion maps for changing data.
\newblock {\em Applied and computational harmonic analysis}, 36(1):79--107,
  2014.

\bibitem{Coifman2006}
R.~R. Coifman and S.~Lafon.
\newblock Diffusion maps.
\newblock {\em Applied and computational harmonic analysis}, 21(1):5--30, 2006.

\bibitem{Coifman2006a}
R.~R. Coifman and S.~Lafon.
\newblock Geometric harmonics: a novel tool for multiscale out-of-sample
  extension of empirical functions.
\newblock {\em Applied and Computational Harmonic Analysis}, 21(1):31--52,
  2006.

\bibitem{coifman2008graph}
R.~R. Coifman, Y.~Shkolnisky, F.~J. Sigworth, and A.~Singer.
\newblock Graph laplacian tomography from unknown random projections.
\newblock {\em IEEE Transactions on Image Processing}, 17(10):1891--1899, 2008.

\bibitem{dauphin2014identifying}
Y.~N. Dauphin, R.~Pascanu, C.~Gulcehre, K.~Cho, S.~Ganguli, and Y.~Bengio.
\newblock Identifying and attacking the saddle point problem in
  high-dimensional non-convex optimization.
\newblock {\em Advances in neural information processing systems}, 27, 2014.

\bibitem{David2012}
G.~David and A.~Averbuch.
\newblock Hierarchical data organization, clustering and denoising via
  localized diffusion folders.
\newblock {\em Applied and Computational Harmonic Analysis}, 33(1):1--23, 2012.

\bibitem{dinh2017sharp}
L.~Dinh, R.~Pascanu, S.~Bengio, and Y.~Bengio.
\newblock Sharp minima can generalize for deep nets.
\newblock {\em Proceedings of the 34th International Conference on Machine
  Learning}, 70:1019--1028, Aug 2017.

\bibitem{Coifman2016}
C.~J. Dsilva, R.~Talmon, C.~W. Gear, R.~R. Coifman, and I.~G. Kevrekidis.
\newblock Data-driven reduction for a class of multiscale fast-slow stochastic
  dynamical systems.
\newblock {\em SIAM Journal on Applied Dynamical Systems}, 15(3):1327--1351,
  2016.

\bibitem{dziugaite2017computing}
G.~K. Dziugaite and D.~M. Roy.
\newblock Computing nonvacuous generalization bounds for deep (stochastic)
  neural networks with many more parameters than training data.
\newblock {\em arXiv preprint arXiv:1703.11008}, 2017.

\bibitem{Farbman:2010}
Z.~Farbman, R.~Fattal, and D.~Lischinski.
\newblock Diffusion maps for edge-aware image editing.
\newblock {\em ACM Transactions on Graphics (TOG)}, 29(6):1--10, 2010.

\bibitem{freeman2016topology}
C.~D. Freeman and J.~Bruna.
\newblock Topology and geometry of half-rectified network optimization.
\newblock {\em arXiv preprint arXiv:1611.01540}, 2016.

\bibitem{garipov2018loss}
T.~Garipov, P.~Izmailov, D.~Podoprikhin, D.~P. Vetrov, and A.~G. Wilson.
\newblock Loss surfaces, mode connectivity, and fast ensembling of dnns.
\newblock {\em Advances in neural information processing systems}, 31, 2018.

\bibitem{Gepshtein:2013}
S.~Gepshtein and Y.~Keller.
\newblock Image completion by diffusion maps and spectral relaxation.
\newblock {\em IEEE Transactions on Image Processing}, 22(8):2983--2994, 2013.

\bibitem{gur2018gradient}
G.~Gur-Ari, D.~A. Roberts, and E.~Dyer.
\newblock Gradient descent happens in a tiny subspace.
\newblock {\em arXiv preprint arXiv:1812.04754}, 2018.

\bibitem{Haddad2014}
A.~Haddad, D.~Kushnir, and R.~R. Coifman.
\newblock Texture separation via a reference set.
\newblock {\em Applied and Computational Harmonic Analysis}, 36(2):335--347,
  2014.

\bibitem{hardt2016train}
M.~Hardt, B.~Recht, and Y.~Singer.
\newblock Train faster, generalize better: Stability of stochastic gradient
  descent.
\newblock {\em Proceedings of the 33rd International Conference on Machine
  Learning}, 48:1225--1234, Jun 2016.

\bibitem{NIPS2019_8524}
H.~He, G.~Huang, and Y.~Yuan.
\newblock Asymmetric valleys: Beyond sharp and flat local minima.
\newblock {\em Advances in neural information processing systems}, 32, 2019.

\bibitem{hochreiter1995simplifying}
S.~Hochreiter and J.~Schmidhuber.
\newblock Simplifying neural nets by discovering flat minima.
\newblock {\em Advances in neural information processing systems}, 7, 1994.

\bibitem{hochreiter1997flat}
S.~Hochreiter and J.~Schmidhuber.
\newblock Flat minima.
\newblock {\em Neural computation}, 9(1):1--42, 1997.

\bibitem{hoffer2017train}
E.~Hoffer, I.~Hubara, and D.~Soudry.
\newblock Train longer, generalize better: closing the generalization gap in
  large batch training of neural networks.
\newblock {\em Advances in neural information processing systems}, 30, 2017.

\bibitem{hu2019diffusion}
W.~Hu, C.~J. Li, L.~Li, and J.-G. Liu.
\newblock On the diffusion approximation of nonconvex stochastic gradient
  descent.
\newblock {\em Annals of Mathematical Sciences and Applications}, 4(1), 2019.

\bibitem{kawaguchi2016deep}
K.~Kawaguchi.
\newblock Deep learning without poor local minima.
\newblock {\em Advances in neural information processing systems}, 29, 2016.

\bibitem{kleinberg2018alternative}
B.~Kleinberg, Y.~Li, and Y.~Yuan.
\newblock An alternative view: When does {SGD} escape local minima?
\newblock {\em Proceedings of the 35th International Conference on Machine
  Learning}, 80:2698--2707, Jul 2018.

\bibitem{Lafon2006}
S.~Lafon, Y.~Keller, and R.~R. Coifman.
\newblock Data fusion and multicue data matching by diffusion maps.
\newblock {\em IEEE Transactions on pattern analysis and machine intelligence},
  28(11):1784--1797, 2006.

\bibitem{LafonThesis}
S.~S. Lafon.
\newblock {\em Diffusion maps and geometric harmonics}.
\newblock PhD thesis, Yale University, 2004.

\bibitem{li2018visualizing}
H.~Li, Z.~Xu, G.~Taylor, C.~Studer, and T.~Goldstein.
\newblock Visualizing the loss landscape of neural nets.
\newblock {\em Advances in neural information processing systems}, 31, 2018.

\bibitem{liang2018understanding}
S.~Liang, R.~Sun, Y.~Li, and R.~Srikant.
\newblock Understanding the loss surface of neural networks for binary
  classification.
\newblock {\em Proceedings of the 35th International Conference on Machine
  Learning}, 80:2835--2843, Jul 2018.

\bibitem{mahalanobis1936generalized}
P.~C. Mahalanobis.
\newblock On the generalized distance in statistics.
\newblock {\em Proceedings of the National Institute of Science of India},
  12:49--55, 1936.

\bibitem{Mishne2013}
G.~Mishne and I.~Cohen.
\newblock Multiscale anomaly detection using diffusion maps.
\newblock {\em IEEE Journal of selected topics in signal processing},
  7(1):111--123, 2012.

\bibitem{neyshabur2017exploring}
B.~Neyshabur, S.~Bhojanapalli, D.~McAllester, and N.~Srebro.
\newblock Exploring generalization in deep learning.
\newblock {\em Advances in neural information processing systems}, 30, 2017.

\bibitem{nguyen2019connected}
Q.~Nguyen.
\newblock On connected sublevel sets in deep learning.
\newblock {\em Proceedings of the 36th International Conference on Machine
  Learning}, pages 4790--4799, 2019.

\bibitem{nguyen2018loss}
Q.~Nguyen, M.~C. Mukkamala, and M.~Hein.
\newblock On the loss landscape of a class of deep neural networks with no bad
  local valleys.
\newblock {\em arXiv preprint arXiv:1809.10749}, 2018.

\bibitem{sagun2017empirical}
L.~Sagun, U.~Evci, V.~U. Guney, Y.~Dauphin, and L.~Bottou.
\newblock Empirical analysis of the hessian of over-parametrized neural
  networks.
\newblock {\em arXiv preprint arXiv:1706.04454}, 2017.

\bibitem{singer2009detecting}
A.~Singer, R.~Erban, I.~G. Kevrekidis, and R.~R. Coifman.
\newblock Detecting intrinsic slow variables in stochastic dynamical systems by
  anisotropic diffusion maps.
\newblock {\em Proceedings of the National Academy of Sciences},
  106(38):16090--16095, 2009.

\bibitem{Singer:2009}
A.~Singer, Y.~Shkolnisky, and B.~Nadler.
\newblock Diffusion interpretation of nonlocal neighborhood filters for signal
  denoising.
\newblock {\em SIAM Journal on Imaging Sciences}, 2(1):118--139, 2009.

\bibitem{smith2017bayesian}
S.~L. Smith and Q.~V. Le.
\newblock A bayesian perspective on generalization and stochastic gradient
  descent.
\newblock {\em arXiv preprint arXiv:1710.06451}, 2017.

\bibitem{Talmon2012}
R.~Talmon, I.~Cohen, and S.~Gannot.
\newblock Single-channel transient interference suppression with diffusion
  maps.
\newblock {\em IEEE transactions on audio, speech, and language processing},
  21(1):132--144, 2012.

\bibitem{tsuzuku2019normalized}
Y.~Tsuzuku, I.~Sato, and M.~Sugiyama.
\newblock Normalized flat minima: Exploring scale invariant definition of flat
  minima for neural networks using {PAC}-{B}ayesian analysis.
\newblock {\em Proceedings of the 37th International Conference on Machine
  Learning}, 119:9636--9647, Jul 2020.

\bibitem{venturi2018spurious}
L.~Venturi, A.~S. Bandeira, and J.~Bruna.
\newblock Spurious valleys in two-layer neural network optimization landscapes.
\newblock {\em arXiv preprint arXiv:1802.06384}, 2018.

\bibitem{wu2020noisy}
J.~Wu, W.~Hu, H.~Xiong, J.~Huan, V.~Braverman, and Z.~Zhu.
\newblock On the noisy gradient descent that generalizes as {SGD}.
\newblock {\em Proceedings of the 37th International Conference on Machine
  Learning}, 119:10367--10376, Jul 2020.

\bibitem{wu2017towards}
L.~Wu, Z.~Zhu, et~al.
\newblock Towards understanding generalization of deep learning: Perspective of
  loss landscapes.
\newblock {\em arXiv preprint arXiv:1706.10239}, 2017.

\bibitem{xie2020artificial}
Z.~Xie, F.~He, S.~Fu, I.~Sato, D.~Tao, and M.~Sugiyama.
\newblock Artificial neural variability for deep learning: on overfitting,
  noise memorization, and catastrophic forgetting.
\newblock {\em Neural computation}, 33(8):2163--2192, 2021.

\bibitem{xie2021covariance}
Z.~Xie, I.~Sato, and M.~Sugiyama.
\newblock A diffusion theory for deep learning dynamics: Stochastic gradient
  descent exponentially favors flat minima.
\newblock {\em arXiv preprint arXiv:2002.03495}, 2020.

\bibitem{xie2020stable}
Z.~Xie, I.~Sato, and M.~Sugiyama.
\newblock Understanding and scheduling weight decay.
\newblock {\em arXiv preprint arXiv:2011.11152}, 2020.

\bibitem{yao2018hessian}
Z.~Yao, A.~Gholami, Q.~Lei, K.~Keutzer, and M.~W. Mahoney.
\newblock Hessian-based analysis of large batch training and robustness to
  adversaries.
\newblock {\em Advances in Neural Information Processing Systems}, 31, 2018.

\bibitem{zhang2017understanding}
C.~Zhang, S.~Bengio, M.~Hardt, B.~Recht, and O.~Vinyals.
\newblock Understanding deep learning (still) requires rethinking
  generalization.
\newblock {\em Communications of the ACM}, 64(3):107--115, 2021.

\bibitem{zhu2018anisotropic}
Z.~Zhu, J.~Wu, B.~Yu, L.~Wu, and J.~Ma.
\newblock The anisotropic noise in stochastic gradient descent: Its behavior of
  escaping from sharp minima and regularization effects.
\newblock {\em arXiv preprint arXiv:1803.00195}, 2018.

\end{thebibliography}

\clearpage 
\appendix
\section{Proof of Equation \ref{eq:sgd_cov_exact}}\label{appendix:proof_cov}
\begin{proof}
First, note that $\nabla \widetilde f^{(k)}(\mathbf{x})$ is an unbiased estimator of $\nabla f(\mathbf{x})$:
\begin{equation*}
\begin{split}
    \mathbb{E}\left[\nabla \widetilde f^{(k)}(\mathbf{x})\right]
    &= \mathbb{E}\left[\frac{1}{n}\sum_{i\in\Omega_k} \nabla f_i(\mathbf{x})\right] = \mathbb{E}\left[\frac{1}{n}\sum_{i=1}^N \nabla f_i(\mathbf{x})\mathds{1}_{i\in\Omega_k}\right] = \frac{1}{n}\sum_{i=1}^N \nabla f_i(\mathbf{x})\mathbb{E}\left[\mathds{1}_{i\in\Omega_k}\right]\\
    &= \frac{1}{n}\sum_{i=1}^N \nabla f_i(\mathbf{x})\frac{n}{N} = \frac{1}{N}\sum_{i=1}^N \nabla f_i(\mathbf{x}) = \nabla f(\mathbf{x}).\\
\end{split}
\end{equation*}

Hence, calculating the covariance,
\begin{equation*}
\begin{split}
    C(x) &= \mathbb{E}\left[\nabla \widetilde f^{(k)}(\mathbf{x}) \nabla \widetilde f^{(k)}(\mathbf{x})^\ast\right] - \mathbb{E}\left[\nabla \widetilde f^{(k)}(\mathbf{x})\right]\mathbb{E}\left[\nabla \widetilde f^{(k)}(\mathbf{x})^\ast\right] \\
    &= \mathbb{E}\left[\left(\frac{1}{n}\sum_{i=1}^N \nabla f_i(\mathbf{x})\mathds{1}_{i\in\Omega_k}\right) \left(\frac{1}{n}\sum_{i=1}^N \nabla f_i(\mathbf{x})\mathds{1}_{i\in\Omega_k}\right)^\ast\right] - \nabla f(\mathbf{x})\nabla f(\mathbf{x})^\ast \\
    &= \frac{1}{n^2} \sum_{i,i^{\prime}=1}^N \nabla f_i(\mathbf{x}) \nabla f_{i^{\prime}}(\mathbf{x})^\ast \mathbb{E}\left[\mathds{1}_{i\in\Omega_k} \mathds{1}_{{i^{\prime}}\in\Omega_k}\right] - \nabla f(\mathbf{x})\nabla f(\mathbf{x})^\ast\\
    &= \frac{1}{n^2} \sum_{i,i^{\prime}=1}^N \nabla f_i(\mathbf{x}) \nabla f_{i^{\prime}}(\mathbf{x})^\ast \mathbb{P}\left(i\in\Omega_k, {i^{\prime}}\in\Omega_k\right) - \nabla f(\mathbf{x})\nabla f(\mathbf{x})^\ast\\
    &= \frac{1}{n^2} \sum_{i,i^{\prime}=1}^N \nabla f_i(\mathbf{x}) \nabla f_{i^{\prime}}(\mathbf{x})^\ast \left[\frac{n}{N}\delta_{ii^{\prime}} - \frac{n}{N}\frac{n-1}{N-1}\left(1 - \delta_{ii^{\prime}}\right)\right] - \nabla f(\mathbf{x})\nabla f(\mathbf{x})^\ast\\
    &= \frac{1}{Nn}\frac{N-n}{N-1} \sum_{i=1}^N \nabla f_i(\mathbf{x}) \nabla f_i(\mathbf{x})^\ast + \left[\frac{1}{Nn}\frac{n-1}{N-1} - \frac{1}{N^2}\right]\sum_{i,i^{\prime}=1}^N \nabla f_i(\mathbf{x}) \nabla f_{i^{\prime}}(\mathbf{x})^\ast\\
    &= \frac{1}{Nn}\frac{N-n}{N-1} \sum_{i=1}^N \nabla f_i(\mathbf{x}) \nabla f_i(\mathbf{x})^\ast + \frac{n-N}{n(N-1)}\frac{1}{N^2}\sum_{i,i^{\prime}=1}^N \nabla f_i(\mathbf{x}) \nabla f_{i^{\prime}}(\mathbf{x})^\ast\\
    &= \frac{1}{Nn}\frac{N-n}{N-1} \sum_{i=1}^N \nabla f_i(\mathbf{x}) \nabla f_i(\mathbf{x})^\ast + \frac{n-N}{n(N-1)}\nabla f(\mathbf{x})\nabla f(\mathbf{x})^\ast\\
    &= \frac{N-n}{n(N-1)}{\Bigg[{\frac{1}{N}}{\sum_{i=1}^N \nabla f_i(\mathbf{x}) \nabla f_i(\mathbf{x})^\ast} -  \nabla f(\mathbf{x}) \nabla f(\mathbf{x})^\ast \Bigg]}.
\end{split}
\end{equation*}
\end{proof}

\end{document}